\documentclass{article}

\usepackage{iclr2027_conference,times}
\usepackage{amsmath,amsfonts,bm}

\def\eqref#1{equation~\ref{#1}}
\def\1{\bm{1}}

\DeclareMathAlphabet{\mathsfit}{\encodingdefault}{\sfdefault}{m}{sl}
\SetMathAlphabet{\mathsfit}{bold}{\encodingdefault}{\sfdefault}{bx}{n}

\usepackage{amsmath,amssymb,mathtools,bm}
\usepackage{nicefrac}

\usepackage{graphicx}
\usepackage{subcaption}
\usepackage{booktabs}
\usepackage{multirow}
\usepackage{tabularx}
\usepackage{array}
\usepackage{placeins}
\usepackage{wrapfig}
\usepackage{makecell}

\usepackage{microtype}
\usepackage{enumitem}
\usepackage{xcolor}
\usepackage{url}
\usepackage{hyperref}
\usepackage[nameinlink,capitalize,noabbrev]{cleveref}

\graphicspath{{figures/}}

\newcommand{\Du}{D_{\mathrm{u}}}
\newcommand{\Dtrain}{D_{\mathrm{train}}}

\definecolor{todopurple}{RGB}{137,103,151}

\newcommand{\fixed}{\textsc{Fixed}}
\renewcommand{\ss}{\textsc{Shuffle}}
\newcommand{\ds}{\textsc{Repack}}

\newcommand{\myskip}[1]{}

\makeatletter
\newcommand*{\myfontsize}{%
  \@setfontsize\myfontsize{8}{9.5}%
}
\makeatother

\let\oldparagraph\paragraph
\renewcommand{\paragraph}[1]{\vspace{-.8em}
\oldparagraph{#1}
}

\title{
    \normalfont Scaling Repeated-Data Pretraining: \\
    Batch Size, Regularization, and GPU-Time Tradeoffs
}

\author{Sewon Min \\ UC Berkeley \& Allen Institute for AI \\\texttt{sewonm@berkeley.edu}}

\iclrfinalcopy
\begin{document}

\maketitle

\begin{abstract}
We study pretraining when compute is abundant but unique training data are finite, requiring repeated training on the same data. We first revisit infinite-compute pretraining~\citep{kim2025infinite} and show that its frontier depends critically on choices previously fixed or studied only in isolation. In particular, optimizing batch size together with epoch counts reveals a U-shaped relationship between validation loss and batch size, favoring smaller batches than typical pretraining, likely because small-batch gradient noise provides implicit regularization. We further find that training recipes, previously optimized under a single-pass assumption, can be tuned, e.g., repacking documents into new sequences every epoch, weight tying, and positive embedding weight decay.

We then relax the infinite-compute assumption. In particular, smaller-than-typical batch sizes can require prohibitively long sequential training, motivating to treat GPUs and time as separate resources.
Tighter deadlines favor larger batches, and latency-oriented parallelism can trade additional GPUs for shorter training. 
Nevertheless, even under a fixed time budget, a smaller batch with fewer GPUs can achieve lower validation loss than a larger batch with more GPUs, challenging the common practice of increasing data parallelism to maximize GPU utilization under a deadline.
Together, our results identify underexplored factors in repeated-data training and provide concrete guidance on how training recipes and resource allocation should change from single-pass pretraining.

\end{abstract}

\section{Introduction}
\label{sec:introduction}

Most language-model pretraining assumes a single pass over a large, diverse web corpus, with compute as the binding resource and data treated as effectively unlimited~\citep{kaplan2020scaling,hoffmann2022training}. However, the supply of high-quality human-written data is finite and grows at a physically limited rate~\citep{villalobos2024data}, so continued scaling increasingly requires training repeatedly on the same data~\citep{muennighoff2023scaling}. Prior work asks how much performance can be extracted from fixed data when compute is effectively unlimited~\citep{kim2025infinite,xu2026dataconstrained}.

\begin{figure*}[t]
    \centering
    \begin{minipage}[b]{0.604\textwidth}
        \includegraphics[width=\linewidth,trim=0 0 204.644635bp 0,clip]{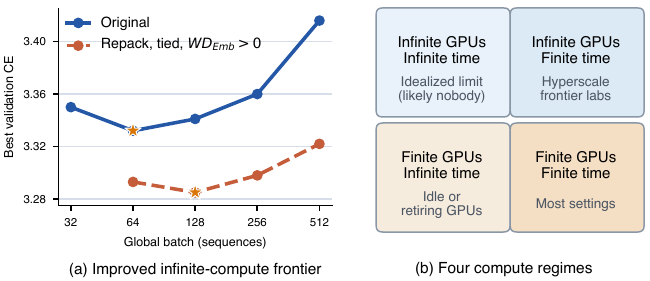}
    \end{minipage}\hfill
    \begin{minipage}[b]{0.392\textwidth}
        \centering
        \includegraphics[width=\linewidth]{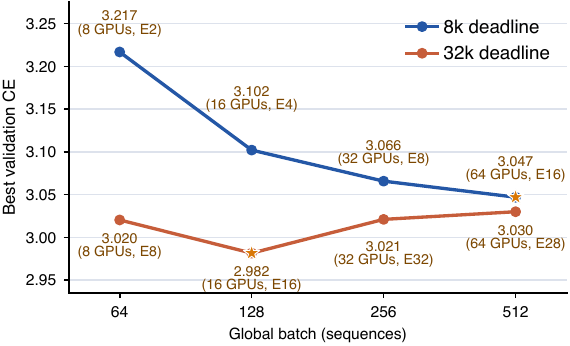}\\[-.1em]
        {\sffamily\fontsize{6}{7}\selectfont (c) Batch size under fixed deadlines}
    \end{minipage}\vspace{-.4em}
    \caption{
    \textbf{(a)} Batch-tuned infinite-compute frontier for a 153M model on the 1B-token data pool. Batch tuning and the repetition-tuned recipe progressively improve the validation loss.
    \textbf{(b)} Separating GPU availability from training time yields four distinct compute regimes.
    \textbf{(c)} An example with a 1.5B model on the 1B-token pool: even when time-constrained, a smaller batch size with fewer GPUs can achieve lower validation cross-entropy (CE) loss than a larger batch size with more GPUs, especially as compute becomes abundant.
    }
    \label{fig:overview}
\end{figure*}

We revisit this setting and show that the infinite-compute frontier depends critically on choices previously fixed or studied only in isolation. In particular, jointly optimizing batch size, learning rate, weight decay, and the number of repetitions\footnote{
    In practice, a training dataset contains repetitions from natural duplication or intentional upsampling of high-quality data. In this work, we simplify the setting and uniformly repeat a fixed dataset.
} reveals a U-shaped relationship between validation loss and batch size, with the optimum at smaller batches than in conventional pretraining. The gradient noise from smaller batches, which can be undesirable in single-pass training, likely provides useful implicit regularization against repeated-data overfitting.
Batch tuning extends beneficial repetitions from 8 to 20 for a 1.5B model and from 40 to 112 for a 153M model, both with substantial loss improvements (Figure~\ref{fig:overview}(a), blue).

We further find that training recipes, which are typically optimized under a single-pass assumption, can be adapted specifically for repeated-data training. We evaluate three simple practices. \ds—repacking documents into new sequences each epoch—creates fresh sequences even when the underlying data remain fixed, acting as a form of repeated-data regularization. Tying the input and output embeddings and applying positive weight decay to the shared parameters provide additional gains. Together, these choices improve validation loss and shift the best checkpoint from epoch 20 to 32 for the 1.5B model and from epoch 112 to 240 for the 153M model (Figure~\ref{fig:overview}(a), blue vs.\ orange).

We next move beyond the infinite compute to consider compute resources, as compute is not truly infinite in practice. In particular, the improved frontier can require prohibitively long sequential training, as it favors smaller-than-typical batch sizes. 
For example, our fitted extrapolation suggests that, for a 70B model trained on 10T unique tokens, reaching the near-optimal stopping point could require over 20 years.
Motivated by this tension, we model GPU availability and sequential training time as separate resource constraints (Figure~\ref{fig:overview}(b)). Under short deadlines, larger batches process more data per optimizer step; as more time becomes available, progressively smaller batches become preferable until the best observed unconstrained $B^\star$ is reached. Separately, latency-oriented parallelism can trade additional GPUs for faster optimizer steps.
Nevertheless, even under a fixed time budget, a smaller batch with fewer GPUs can achieve lower validation loss than a larger batch with more GPUs, especially with longer deadlines (Figure~\ref{fig:overview}(c)). This finding suggests that the common practice of maximizing parallelism with more GPUs under a fixed deadline may be suboptimal for repeated-data training.

In summary, this paper shows that, when unique data become the binding constraint, both pretraining recipes and compute allocation must change.
\begin{enumerate}[leftmargin=*,topsep=2pt,itemsep=2pt]
    \item \textbf{Batch size must account for repeated-data generalization.} Repeated-data training favors smaller batches than those used in conventional pretraining (\S\ref{sec:batch_size_two_dimensional}).
    \item \textbf{Training recipes must be adapted to repeated exposure.} Repacking documents across epochs and regularizing embeddings improve attainable loss and make more repetitions beneficial (\S\ref{sec:repetition_aware}).
    \item  \textbf{Compute allocation must distinguish GPUs from time.} Maximizing GPU utilization through larger batches can be suboptimal: even under the same deadline, smaller batches using fewer GPUs can achieve lower loss  (\S\ref{sec:unified_gpu_time_scaling}).
    %\item We show that repeated-data training has its own statistical batch-size optimum (\S\ref{sec:batch_size_two_dimensional}), which also motivates decomposing compute into GPUs and time.
    %\item We show that the training recipe itself can be adapted to repeated-data training by varying how data are presented across repetitions and regularizing the embeddings (\S\ref{sec:repetition_aware}).
    %\item Finally, we map the resulting optimal recipes to the available GPUs and time, showing that (1) although the unconstrained optimum can be very slow, we can jointly optimize repetitions, batch size, and compute configuration under GPU and time budgets, and (2) it can often overturn conventional decisions (\S\ref{sec:unified_gpu_time_scaling}).
\end{enumerate}

\section{Related Work}
\label{sec:related_work}

\oldparagraph{Compute-bound scaling.}
Conventional language-model scaling laws predict loss from model size, training tokens, and compute, then use these relationships to select compute-optimal model--data allocations~\citep{kaplan2020scaling,hoffmann2022training}. 
They effectively assume an unlimited supply of fresh data, likely because available compute at the time had not yet reached the scale needed to exhaust the available stock of human-generated web data.

\paragraph{Data-bound scaling.}
As hardware improves and the stock of high-quality human-written text grows slowly, unique data increasingly becomes a binding resource~\citep{villalobos2024data}. This has motivated repeated-data pretraining, in which models train on a fixed dataset until validation gains saturate or reverse because of overfitting~\citep{muennighoff2023scaling,xue2023repeat}. At fixed compute, \citet{fang2025datasets} show that repeatedly training on filtered data can outperform training on a $10\times$ larger superset with proper regularization, e.g., weight decay tuning.

The infinite-compute, finite-data setting is studied most directly by two recent papers. \citet{kim2025infinite} fix the unique data, remove the compute constraint, and show that stronger regularization, especially weight decay tuning, substantially delays overfitting. \citet{xu2026dataconstrained} build on this work and propose additional regularization methods and an interaction-aware scaling law. We revisit the same setting but optimize previously fixed recipe dimensions and then ask how the resulting statistical frontier changes under separate GPU and time constraints.

\paragraph{Batch size.}
Batch size optimization in conventional single-pass pretraining balances statistical efficiency and parallelism.
At a fixed number of optimizer steps, larger batches process more tokens and are generally favored; at a fixed token budget, smaller batches take more steps and can be more token-efficient.
Prior work identifies a critical batch size below which jointly increasing batch size and learning rate preserves token efficiency while improving throughput, and above which token efficiency degrades~\citep{mccandlish2018empirical,kaplan2020scaling,zhang2024critical,merrill2025critical}.
However, reduced token efficiency alone does not imply worse loss under a fixed deadline, since larger batches process more tokens in the available time. Therefore, conventional recipes often increase batch size up to the critical batch size to expose parallelism. 

For repeated-data training, batch size has  received less attention. \citet{kim2025infinite} select a small batch using a single-pass ablation and then hold it fixed in their repeated-data scaling experiments; \citet{xu2026dataconstrained} likewise use a fixed batch throughout.

We instead fix the unique-data pool and jointly tune the learning rate, weight decay, and repetitions for each batch size. We find that repeated-data training changes the conventional tradeoff: smaller batches can achieve lower validation loss, even under a fixed deadline.

\section{Batch Size Shapes the Repeated-Data Frontier}
\label{sec:batch_size_two_dimensional}

We first revisit the unconstrained repeated-data frontier by jointly optimizing batch size, learning rate, weight decay, and epoch count, and reveal an interior batch-size optimum. Because batch size controls both statistical behavior and available parallelism, this later motivates separating GPU and time constraints in \S\ref{sec:unified_gpu_time_scaling}.

\subsection{Experimental Setup}
\label{subsec:batch_tuning_setup}

\oldparagraph{Models \& data.}
We train dense decoder-only Transformers with 153M, 474M, and 1.5B parameters, using the \href{https://github.com/allenai/OLMo-core}{olmo-core}~\citep{groeneveld2024olmo} codebase. % using the public \href{https://github.com/allenai/EMO}{\texttt{EMO}} codebase~\citep{wang2026emo}, built on the \href{https://github.com/allenai/OLMo-core}{olmo-core}~\citep{groeneveld2024olmo}.
All experiments use DCLM-Baseline~\citep{li2024datacomplm} as the training corpus, with a 4,096-token context and document-disjoint training, validation, and test splits. This section uses a 1B-token unique-data pool; \S\ref{subsec:recipe_interventions_results} expands the study to 0.333B-, 1B-, and 3B-token pools.
Together, the measured grid covers $\Du/N$ from 0.22 to 19.61. %, so the frontier-motivated regime $\Du/N>100$ remains an extrapolation target.
We train with AdamW~\citep{loshchilov2019adamw} and a warmup-stable-decay (WSD) learning-rate schedule~\citep{hu2024minicpm,wen2024wsd}, with a 100M-token warmup and 10\% terminal decay.
%token-matched warmup and a target-specific 10\% terminal decay.

\paragraph{Hyperparameter tuning methodology.}
For a model with $N$ parameters and $\Du$ unique tokens, $E$ passes yield $\Dtrain=E\Du$ token presentations. % and $S=\Dtrain/B$ sequential optimizer steps at global batch $B$.
We sweep global batches of 16--512 sequences for the 1.5B model and 32--512 for the smaller models.

For each $(N,\Du,B)$, we tune the peak learning rate and weight decay through gradual adaptive coordinate search: we first select the pair for an earlier epoch, then proceed to the next epoch by starting from the pair selected for the previous epoch and including neighboring learning rates ($1/2\times$ and $2\times$) and a weight-decay value one level higher ($3\times$).

We seek the best training recipe at a fixed model size, rather than allowing parameter scaling or ensembling as an alternative route to lower loss~\citep{kim2025infinite}. This choice reflects the practical importance of inference cost.

\paragraph{Evaluation.}
Our primary metric is next-token cross-entropy (CE) on a fixed 100M-token DCLM-Baseline validation split. 

Evaluations on downstream tasks, sensitivity to random seeds, and comparisons with unique-data pools are reported in \S\ref{app:additional_infinite_time_results}.

\begin{figure*}[t]
    \centering
    \includegraphics[width=0.98\textwidth]{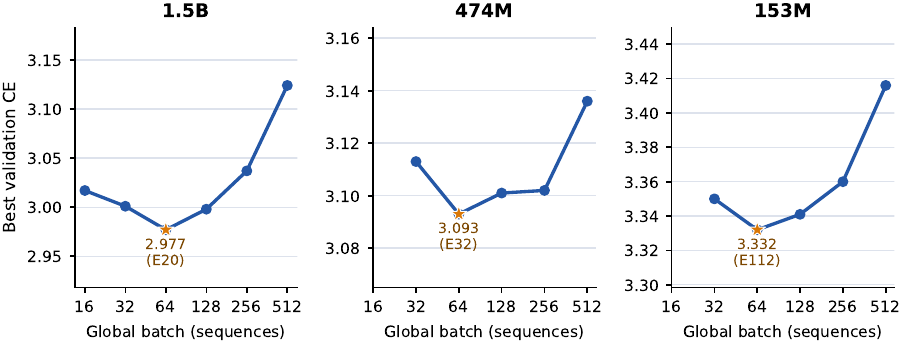}
    \caption{\textbf{Batch size has an interior optimum on the 1B-token pool.} At each (model size, batch size) pair, we tune (learning rate, weight decay, epoch count), and report the lowest validation CE.
    }
    \label{fig:batch_optima_1b_pool}
\end{figure*}

\paragraph{Comparison with prior setups.}
\citet{kim2025infinite} select a small batch using a single-pass ablation and then hold it fixed in their repeated-data scaling experiments; \citet{xu2026dataconstrained} likewise use a fixed batch throughout. We instead optimize batch size specifically for repeated-data training by comparing each candidate over its own training horizon. Other setup differences include sequence order and learning-rate scheduling: Kim et al. reuse a fixed sequence order across epochs and train each horizon with an independent cosine schedule, whereas we reshuffle sequence indices each epoch and use WSD. Xu et al. also use WSD and reshuffle each epoch, making their setting closer to ours. To reduce evaluation noise, we use a fixed 100M-token validation split, compared with 4M tokens in \citet{kim2025infinite} and 10M tokens in \citet{xu2026dataconstrained}.

\subsection{Results}
\label{subsec:batch_tuning_results}

Unless stated otherwise, each point is the best endpoint after tuning learning rate, weight decay, and epoch count.

\paragraph{Batch size has an interior optimum.}
Figure~\ref{fig:batch_optima_1b_pool} shows that batch-size tuning has a substantial effect. Relative to the best $B=512$ run, the best $B=64$ run reduces validation CE from $3.124$ to $2.977$ for the 1.5B model, from $3.136$ to $3.093$ for the 474M model, and from $3.416$ to $3.332$ for the 153M model. For all three model sizes, the complete sweep is U-shaped: validation CE is minimized near $B=64$, with both larger and smaller batches performing worse.

This optimum is notable because $B=64$ is substantially smaller than the batches used in public pretraining recipes at a similar scale: GPT-2, Cerebras-GPT, Pythia, and OLMo use global batches ranging from 512 to 2,048 sequences~\citep{radford2019language,dey2023cerebras,biderman2023pythia,groeneveld2024olmo}.

\begin{wraptable}{r}{0.41\textwidth}
\vspace{-1.5\baselineskip}
\centering
\myfontsize
\setlength{\tabcolsep}{4.5pt}
\caption{\textbf{Batch-size comparison for the 1.5B model on the 1B-token pool.} $C_{\mathrm{F}}\approx6NE\Du$ denotes compute consumed; batches are in millions of tokens.}
\label{tab:compute_optimal_batch_comparison}
\vspace{-.5em}
\begin{tabular}{@{}ll @{\hspace{-1em}} r@{}}
\toprule
& \makecell[c]{$C_{\mathrm{F}}$ (equiv. $E$)} & \makecell[r]{Batch size \\ (million tokens)} \\
\midrule
Bi et al.    & $0.90\times10^{19}$ ($E=1$)  & 0.463 \\
Bi et al.    & $7.20\times10^{19}$ ($E=8$)  & 0.914 \\
Bi et al.    & $18.0\times10^{19}$ ($E=20$) & 1.233 \\
\midrule
Shuai et al. & $0.90\times10^{19}$ ($E=1$)  & 0.551 \\
Shuai et al. & $7.20\times10^{19}$ ($E=8$)  & 0.681 \\
Shuai et al. & $18.0\times10^{19}$ ($E=20$) & 0.747 \\
\midrule
Ours & $\infty$ ($C_{\mathrm{F}}$: $E=20$) & 0.262 \\
\bottomrule
\end{tabular}
\vspace{-1.5\baselineskip}
\end{wraptable}
The gap remains when compared with predictions from more recent research on batch-size scaling.
For comparison, the single-pass laws of \citet{bi2024deepseek} and \citet{shuai2024batch} predict $B_{\mathrm{opt}}=0.2920C_{\mathrm{F}}^{0.3271}$ and $B_{\mathrm{opt}}=6.42\times10^3C_{\mathrm{F}}^{0.102}$, respectively, with $C_{\mathrm{F}}$ in FLOPs and $B$ in tokens.
Our 1.5B model reaches its best observed repeated-data loss after 20 passes over the 1B-token pool, which is equivalent to $1.8\times10^{20}$ FLOPs.
At this compute scale, the two laws predict batches of 1.233M and 0.747M tokens, respectively---4.7 and 2.9 times our selected batch of 0.262M tokens (Table~\ref{tab:compute_optimal_batch_comparison}). This suggests that the batch size derived from single-pass training may be suboptimal for repeated-data regime, and the repeated-data optimum can be substantially smaller.

\begin{figure}[ht]
    \centering
    \vspace{-.5em}
    \includegraphics[width=.9\linewidth]{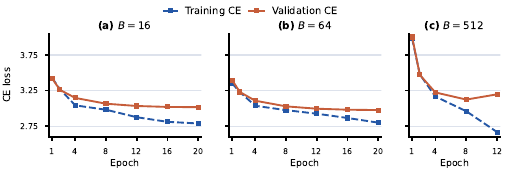}
    % \vspace{-1em}
    \caption{\textbf{Training and validation losses across batch sizes.} The large batch ($B=512$) sees a larger train-validation gap and earlier validation degradation, consistent with weaker stochastic regularization.}
    \label{fig:train_validation_dynamics}
    \vspace{-1em}
\end{figure}

\paragraph{Optimization and generalization jointly shape the U-curve.}
Then, why can't batch size be too large or too small?
Figure~\ref{fig:train_validation_dynamics} compares training loss, validation loss, and their gap across batch sizes for the 1.5B model.
\begin{itemize}[leftmargin=14pt, topsep=1pt,itemsep=0pt]
    \item $B=512$ vs. $B=64$: Early in training, $B=512$ is under-optimized because it takes fewer optimizer steps per epoch. Later, however, its training loss drops below that of $B=64$, so under-optimization can no longer explain the difference. Its validation loss still remains worse and eventually increases. This pattern is consistent with weaker stochastic regularization at large batch sizes~\citep{keskar2017largebatch,smith2020noise}.
    \item $B=16$ vs. $B=64$: Their training losses are similar early in training. At later epochs, $B=16$ achieves lower training loss but higher validation loss than $B=64$. Thus, below $B=64$, the additional optimizer steps eventually lead to overfitting: they continue to reduce training loss without improving held-out performance.
\end{itemize}
In short, the U-curve reflects a balance between optimization and generalization. Relative to conventional large batches, smaller batches benefit from stronger stochastic regularization and delay overfitting; below the optimum, additional optimizer steps stop improving the validation loss.

\paragraph{The importance of weight-decay tuning depends on batch size.}
Prior finite-data studies emphasize tuning weight decay to delay overfitting~\citep{kim2025infinite,fang2025datasets,xu2026dataconstrained}. We likewise find it important for large batches (e.g., $B=512$) and in more severely data-constrained settings. Near the batch optimum, however, its effect is smaller: at $B=64$, the weight decay selected at epoch 1, $0.3$, remains optimal through saturation.

\section{Training Choices Shift the Repeated-Data Frontier}
\label{sec:repetition_aware}

In addition to batch size, we show that recipe details, typically optimized for single-pass training or chosen incidentally, can be tuned specifically for repeated-data training. We aim to improve the frontier across batch sizes rather than for the optimal batch size only, since we later show that larger-than-optimal batches can be preferred under time constraints (\S\ref{sec:unified_gpu_time_scaling}).

\subsection{Controlled Recipe Ablations}
\label{subsec:recipe_interventions}
Unless noted otherwise, all comparisons use the 1.5B model, a 1B-token unique-data pool, sequence length 4,096, WSD, and the tuning methodology from \S\ref{sec:batch_size_two_dimensional}.

\paragraph{Data presentation across epochs.}
When a finite data is repeated, the data loader determines whether the model repeatedly encounters identical packed sequences or new combinations of documents and contexts. We compare three methods with progressively more cross-epoch variation:
\begin{enumerate}[leftmargin=14pt, topsep=1pt,itemsep=0pt]
    \item \fixed: keeping packed sequences, minibatch composition, and minibatch order fixed across epochs, as in \citet{kim2025infinite};
    \item \ss: keeping sequence boundaries fixed while reshuffling sequences across minibatches, producing fresh batches each epoch; and
    \item \ds: reshuffling documents and dynamically repacking them into fixed-length sequences, producing both fresh sequences and fresh batches each epoch.
\end{enumerate}
These methods are equivalent in a single pass but expose models to different examples once data are repeated. Prior repeated-data studies use either \fixed~\citep{kim2025infinite} or fixed minibatch composition with shuffled minibatch order~\citep{fang2025datasets,xu2026dataconstrained}. We test whether the additional variation introduced by \ds\ acts as repeated-data regularization.

\begin{figure}[t]
    \centering \vspace{-1em}
    \footnotesize
    \begin{minipage}[c]{0.48\textwidth}
        \vspace{-10mm}
        \centering
        \textbf{(a) Best validation CE}\par\vspace{1mm}
        \setlength{\tabcolsep}{6pt}
        \begin{tabular}{lrr}
            \toprule
            Loader & $B=64$ & $B=512$ \\
            \midrule
            \fixed & 3.011\,{\scriptsize(E16)} & 3.205\,{\scriptsize(E8)} \\
            \ss & 2.977\,{\scriptsize(E20)} & 3.124\,{\scriptsize(E8)} \\
            \ds & \textbf{2.971\,{\scriptsize(E24)}} & \textbf{3.064\,{\scriptsize(E24)}} \\
            \bottomrule
        \end{tabular}
    \end{minipage}\hfill
    \begin{minipage}[c]{0.48\textwidth}
        \vspace{0pt}
        \centering
        \small
        \textbf{(b) Validation-selected envelopes for $B=512$}\par\vspace{1mm}
        \includegraphics[width=\linewidth,trim=13pt 10pt 2pt 16pt,clip]{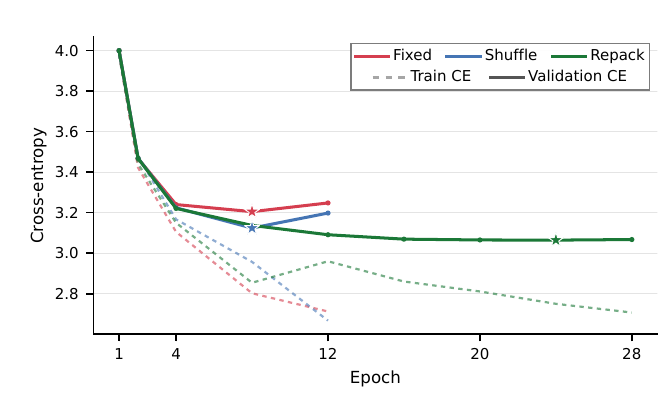}
    \end{minipage}
    \vspace{-3mm}
    \caption{\textbf{Data-loader ablation.} (a) Best validation loss and its selected epoch for each batch and loader, with learning rate and weight decay tuned for each. (b) Validation and train losses for $B=512$; stars indicate each method's best observed validation checkpoint.}
    \label{fig:data_loader_ablation}
\end{figure}
\begin{figure}[t]
    \centering \vspace{-1em}
    \footnotesize
    \begin{minipage}[c]{0.52\textwidth}
        \vspace{-4mm}
        \centering
        \textbf{(a) Best validation CE}\par\vspace{1mm}
        \setlength{\tabcolsep}{3pt}
        {\myfontsize %\footnotesize
        \begin{tabular}{llrr}
            \toprule
            Loader & Method & $B=64$ & $B=512$ \\
            \midrule
            \ss & Untied & 2.977\,{\scriptsize(E20)} & 3.124\,{\scriptsize(E8)} \\
            \ds & Untied & 2.971\,{\scriptsize(E24)} & 3.064\,{\scriptsize(E24)} \\
            \ds & Tied, WD$_{\mathrm{Emb}}=0$ & 2.971\,{\scriptsize(E24)} & 3.061\,{\scriptsize(E20)} \\
            \ds & Untied, WD$_{\mathrm{Emb}}>0$ & 2.951\,{\scriptsize(E32)} & 3.050\,{\scriptsize(E28)} \\
            \ds & Tied, WD$_{\mathrm{Emb}}>0$ & \textbf{2.946\,{\scriptsize(E32)}} & \textbf{3.030\,{\scriptsize(E28)}} \\
            \bottomrule
        \end{tabular}}
    \end{minipage}
    \hfill
    \begin{minipage}[c]{0.46\textwidth}
        \vspace{0pt}
        \centering
        \small
        \textbf{(b) Validation-selected envelopes for $B=64$}\par\vspace{1mm}
        \includegraphics[width=\linewidth,trim=13pt 10pt 2pt 16pt,clip]{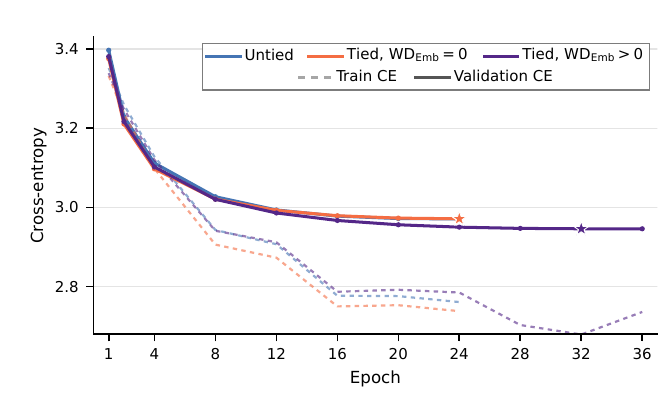}
    \end{minipage}
    \vspace{-3mm}
    \caption{\textbf{Weight-tying and embedding-decay ablation with \ds.} (a) Best validation loss and its selected epoch for each batch size and method, with learning rate and weight decay tuned for each. (b) Validation and train losses for $B=64$; stars indicate each recipe's best validation checkpoint.}
    \label{fig:weight_tying_ablation}
\end{figure}

Figure~\ref{fig:data_loader_ablation} shows results for a 1.5B model with $B=\{64, 512\}$. \fixed\ clearly underperforms \ss\ in validation CE and exhibits a larger train-validation gap at $B=512$.
\ds\ further improves on \ss\ at both batch sizes and reaches its best observed validation loss later in training, with a clearer benefit at $B=512$ than at $B=64$.
These results indicate that, for a fixed unique-data pool, how examples are presented across epochs materially affects the attainable frontier.

\paragraph{Tied input/output embeddings and embedding decay.}
Weight tying shares one parameter matrix between the input embeddings and output head and was originally proposed for parameter reduction~\citep{press-wolf-2017-using}. Although larger contemporary models often leave these matrices untied, compact models continue to use tying, e.g., Qwen3.5-0.8B~\citep{qwen35}.
We ask whether tying and decay on the shared embedding affect repeated-data training. We fix the dataloder to \ds\ and evaluate \{untied model, tied model\} $\times$ \{no embedding decay, positive embedding decay\}. 

Figure~\ref{fig:weight_tying_ablation}(a) shows that weight tying alone yields little to no improvement, while embedding decay consistently improves validation loss.
Combining weight tying and positive embedding decay yields the best results, improving the validation loss from from $2.971$ to $2.946$ at $B=64$ and from $3.064$ to $3.030$ at $B=512$, relative to the original setup of untied embeddings and no embedding decay. It also reaches its best validation loss later, shifting the selected epoch from 24 to 32 and 28, respectively.\footnote{Weight tying reduces the 153M, 474M, and 1.5B models to approximately 102M, 372M, and 1.3B parameters, respectively; we retain the original names for consistency. Because the untied models contain more parameters, this difference does not explain the tied configuration's lower validation loss.}

\paragraph{Failed attempts \& higher-risk recipes (\S\ref{app:higher_risk_improvements}).}
We also tried numerous other approaches, specifically for recovering the regularization benefits of noisy, small-batch gradients while retaining the data parallelism of large-batch training. Methods we tried include batch warmdown, structured gradient noise, and periodically synchronized local optimization methods such as LocalSGD~\citep{stich2019local} and DiLoCo~\citep{douillard2023diloco}.
Discussions and selected results are provided in \S\ref{app:higher_risk_improvements}.
In short, although some of these approaches showed promising signals, none of them yielded improvements as clear or consistent as the methods described in this section, and we therefore did not use them for the rest of our experiments.

\subsection{Complete Results with the Repetition-Tuned Recipe}
\label{subsec:recipe_interventions_results}

\begin{figure*}[t]
    \centering
    \includegraphics[width=\textwidth]{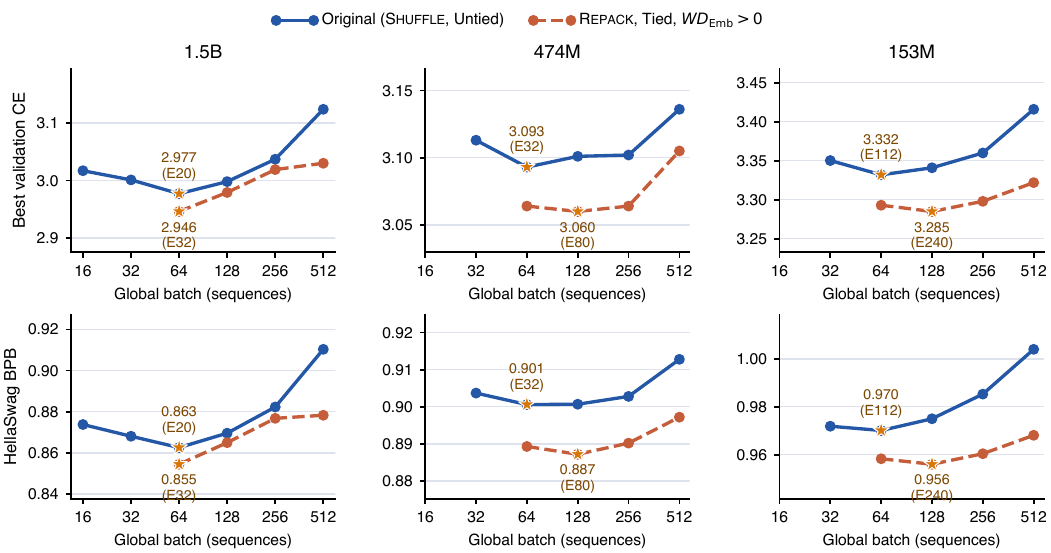}
    \vspace{-1.5em}
    \caption{\textbf{Comparison between the original and repetition-tuned recipes (\ds, tied embeddings, positive embedding decay).} The repetition-tuned recipe improves both validation CE (top) and HellaSwag BPB (bottom) across model sizes and batch sizes.}
    \label{fig:recipe_batch_optima}
\end{figure*}

\oldparagraph{Complete original vs. repetition-aware recipe comparison.} Figure~\ref{fig:recipe_batch_optima} reports a complete comparison between the original (\ss, untied embeddings, and no embedding decay) and repetition-aware (\ds, tied embeddings, and positive embedding decay) recipes across model sizes and batch sizes.
The repetition-aware recipe improves validation loss and HellaSwag bits-per-byte (BPB) across all three model sizes and extends the validation-selected training horizon. The selected epoch moves from 20 to 32 for the 1.5B model, from 32 to 80 for the 474M model, and from 112 to 240 for the 153M model. The recipe also shifts the optimal batch size from 64 to 128 for the 474M and 153M models. It is likely because the additional regularization reduces the marginal regularization benefit of small-batch gradient noise.

\paragraph{Comparison with open-source default.}
Our survey of nine representative open-source training stacks (Table~\ref{tab:open_source_defaults} in \S\ref{app:comparison-with-open-source-defaults}) shows that these ingredients already appear in some stacks, but in inconsistent combinations that often differ from our best-performing configuration and are generally not motivated by repeated-data behavior. Since our experiments show that these choices materially affect attainable loss, we recommend that researchers and LLM developers treat these choices as explicit design decisions rather than blindly inheriting a stack's defaults.

\paragraph{Applying the finalized recipe across data-pool sizes.}
Having established the repetition-aware recipe, we fix it for the rest of the experiments and extend our analysis to three unique-data pools: 0.333B, 1B, and 3B tokens.
We tune the learning rate on the 1B-token pool and transfer it to the 0.333B- and 3B-token pools, following prior evidence that the optimal learning rate is relatively stable across data-pool sizes~\citep{kim2025infinite,xu2026dataconstrained}; our preliminary experiments show the same pattern. We also tune weight decay with the assumption that larger data pools use weight decay values no higher than those used for smaller pools. 

%For the 0.333B-token pool, we evaluate checkpoints every 4, 8, and 16 epochs for the 1.5B, 474M, and 153M models, respectively. For the 3B-token pool, we evaluate every 8, 16, and 32 epochs, respectively, increasing the interval to 64 epochs for the 153M model after epoch 256.

\begin{figure*}[t]
    \centering
    \includegraphics[width=\textwidth]{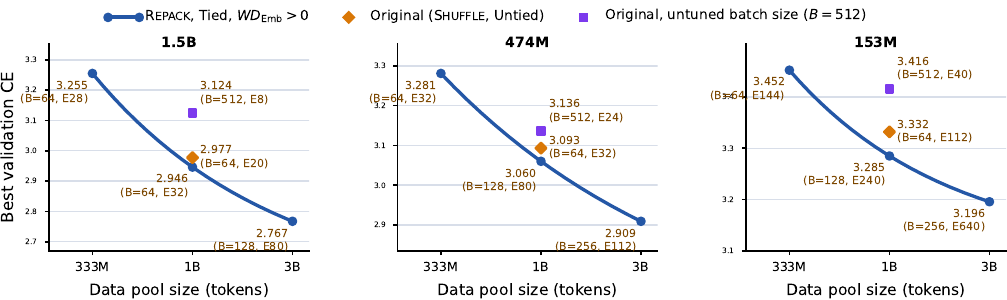}
    \caption{\textbf{The repetition-tuned recipe improves as the unique-data pool grows.} We report the best observed validation CE on 0.333B-, 1B-, and 3B-token pools. Orange diamonds show the tuned original recipe on the 1B-token pool, and purple squares show its untuned $B=512$ baseline. Labels give the selected batch size and epoch.}
    \label{fig:pool_size_best_validation_ce}
\end{figure*}

Figure~\ref{fig:pool_size_best_validation_ce} shows that more unique data improves the best observed validation loss and shifts the selected configuration toward larger batches and later epochs. This is likely because larger pools are less data-constrained: overfitting begins later and additional repetitions remain useful for longer.

\section{Scaling Across GPU and Time Budgets}
\label{sec:unified_gpu_time_scaling}

We now relax the infinite-compute assumption and consider realistic resource constraints.
In particular, our results in \S\ref{sec:batch_size_two_dimensional} motivate treating GPUs and time separately: smaller-than-typical batches may use fewer GPUs but require longer training. 
This matters because some labs may have abundant (nearly unlimited) GPUs but still finite time.
We therefore treat GPU availability and training time as separate constraints and study scaling under limits on time, GPUs, or both.

\subsection{Motivation: Why Separate GPUs and Time?}
\label{subsec:motivation}
To quantify the potential resource implications,
we consider a hypothetical scenario with a dense model with $N=70$B trained on $\Du=10$T unique tokens.
We fit the loss curve for extrapolation and define $E^{\star}$ as the earliest epoch whose fitted validation CE is within 0.001 of its fitted minimum (about 0.1\% in perplexity). We also estimate the corresponding batch size separately.
The fitting procedure and extrapolation are detailed in \S\ref{app:loss_fitting_scaling_law}. Note that this example is an extreme extrapolation far beyond our measured model and data scales, with no target-scale statistical validation; we therefore treat the resulting numbers as illustrative rather than conclusive.

The fit gives $E^{\star}\approx214.2$ and $B^{\star}\approx1{,}536$. At sequence length 4{,}096, this corresponds to approximately 340.1 million optimizer steps. With gradient accumulation $a=1$, microbatch size $m=1$, tensor parallelism $t=4$, pipeline parallelism $p=1$, and the optimistic optimizer-step latency $\tau(70)\approx2.10$ seconds,\footnote{The systems assumptions are detailed in \S\ref{app:parallelism-estimation}.} the corresponding resource requirements are
\[
G^{\star} \approx 6{,}144\ \text{H100 GPUs}
\qquad \qquad
T^{\star} \approx 22.6\ \text{years}.
\]
This extrapolated near-optimal operating point motivates the four resource regimes in Figure~\ref{fig:overview}(b).

\vspace{.1em}\noindent\textbf{1) Infinite GPUs, infinite time.}
If a lab has sufficient GPUs and time, it can execute the infinite-compute recipe directly.

\vspace{.1em}\noindent\textbf{2) Finite GPUs, infinite time.}
With fewer GPUs but sufficient time, gradient accumulation can trade parallelism for sequential execution. The same statistical recipe can therefore still be realized, though more slowly.

\vspace{.1em}\noindent\textbf{3) Infinite GPUs, finite time.}
If GPU availability is effectively nonbinding but training faces a deadline---as may be the case for hyperscale frontier labs---then sequential time becomes the primary constraint. 
Publicly reported training scales suggest that this regime is already plausible: Llama~4 reports using 32,000 GPUs, and Grok~3 reportedly used approximately 100,000 H100s~\citep{meta2025llama4,shilov2025grok3}. Thus, our extrapolated requirement of 6,144 H100s may be attainable for large AI labs; however, the projected 22.6-year training time would likely remain a constraint.

In this regime, the unconstrained recipe generally cannot be used as is. Additional GPUs can shorten training by increasing the data-parallel batch or using more aggressive tensor or pipeline parallelism, but neither tradeoff is free. A larger batch can worsen attainable validation loss, and model parallelism incurs communication and scheduling overhead. This motivates time-constrained scaling in \S\ref{subsec:time_constrained_scaling}. %, jointly selecting the batch size and compute configuration for a given deadline.

\vspace{.1em}\noindent\textbf{4) Finite GPUs, finite time.}
When both resources bind, as is likely in most settings, the stopping epoch, batch size, and compute configuration must be selected jointly for the available budgets. We study this joint GPU-time-constrained regime in \S\ref{subsec:time_gpu_constrained}.

\subsection{Time-Constrained Scaling}
\label{subsec:time_constrained_scaling}

\begin{figure*}[t]
\centering
\includegraphics[width=\textwidth]{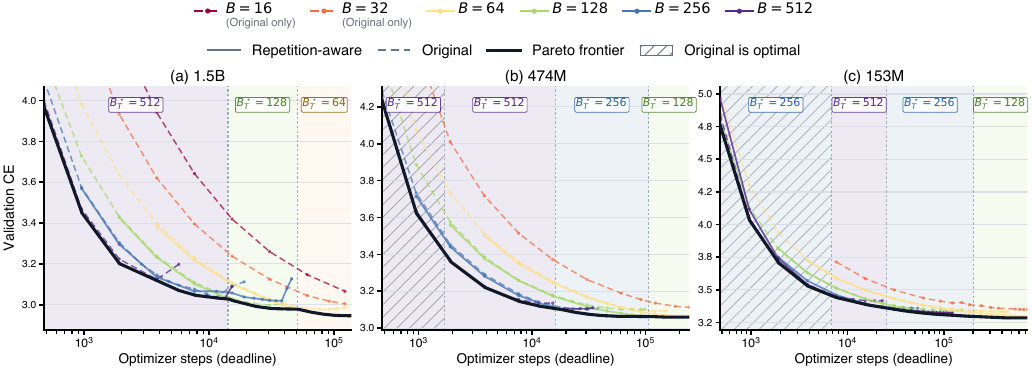}
\caption{\textbf{The time-constrained frontier jointly selects the training recipe, batch size, and the number of repetitions on the 1B-token data pool.} The black curve shows the best validation CE
attainable at each optimizer-step deadline; $B_T^\star$ indicates the batch that supplies the frontier.}
\label{fig:recipe_validation_time_frontier}
\end{figure*}

When training is deadline-constrained, three coupled choices determine the attainable loss:
\begin{enumerate}[leftmargin=14pt, topsep=1pt,itemsep=0pt]
    \item the number of data repetitions (equivalently, the stopping epoch $E$),
    \item the global batch size $B$, and
    \item the compute-parallelism configuration, including the microbatch size and the DP, TP, and PP degrees.
\end{enumerate}
Stopping early while retaining the unconstrained batch $B^\star$ may expose the model to too little data. On the other hand, increasing the batch processes more data per optimizer step, but can eventually yield worse validation loss as repeated-data overfitting emerges. This motivates a deadline-aware recipe that jointly optimizes these choices.

\paragraph{Short deadlines favor larger batches.} Figure~\ref{fig:recipe_validation_time_frontier} shows this tradeoff systematically across training recipes, epoch counts $E$, and batch sizes $B$, across all three model sizes.
Under short budgets, larger batches tend to be preferred because they process more data per optimizer step. As the budget increases, progressively smaller batches become preferable, until the frontier reaches the best observed unconstrained batch $B^\star$. 

Importantly, a fixed time budget does not imply that larger batches are always preferable. Figure~\ref{fig:overview}(c) illustrates this for the 1.5B model. Under a short deadline (equivalent to 8K steps), a larger batch using 64 GPUs achieves the best loss. As the deadline increases (equivalent to 32K steps), however, a small batch using only 16 GPUs achieves a lower loss, despite using strictly fewer GPUs. This indicates that maximizing available GPUs by increasing the global batch size can be suboptimal.

\paragraph{Deadlines also change the preferred training recipe.} For small models (474M and 153M), \textsc{Original} may be preferred under very short deadlines, likely because repeated-data overfitting has not yet become relevant. The repetition-aware recipe subsequently becomes optimal and remains so at longer deadlines, including across the entire observed horizon for the 1.5B model. The difference is consistent but minor, likely because it primarily helps by delaying overfitting and increasing the number of beneficial repetitions.

\paragraph{Parallelism can instead be spent on latency.}
Batch size is not the only way to trade compute for training time.
The compute-parallelism configuration itself can be optimized for latency rather than throughput, using additional GPUs to shorten training without changing the global batch size, e.g., by minimizing the microbatch size and using more aggressive model parallelism. \S\ref{subsec:aggressive_model_parallelism} reports projections from a simulated, communication-aware H100 model; these are illustrative rather than end-to-end measurements. In this model, reducing the microbatch size from the throughput-optimal value of 8 to 1 uses $8\times$ more GPUs and shortens 1.5B-model training by $2.7\times$ at unchanged validation CE. At the extrapolated 70B scale, using $8\times$ more model parallelism yields a projected $3.75\times$ speedup; additional model parallelism provides no further modeled latency reduction. We do not include these projections in figures and the subsequent \S\ref{subsec:time_gpu_constrained} as we focus on the empirically observed frontier, but recommend future work to explore this direction.

\begin{figure*}[t]
\centering
\includegraphics[width=.9\textwidth]{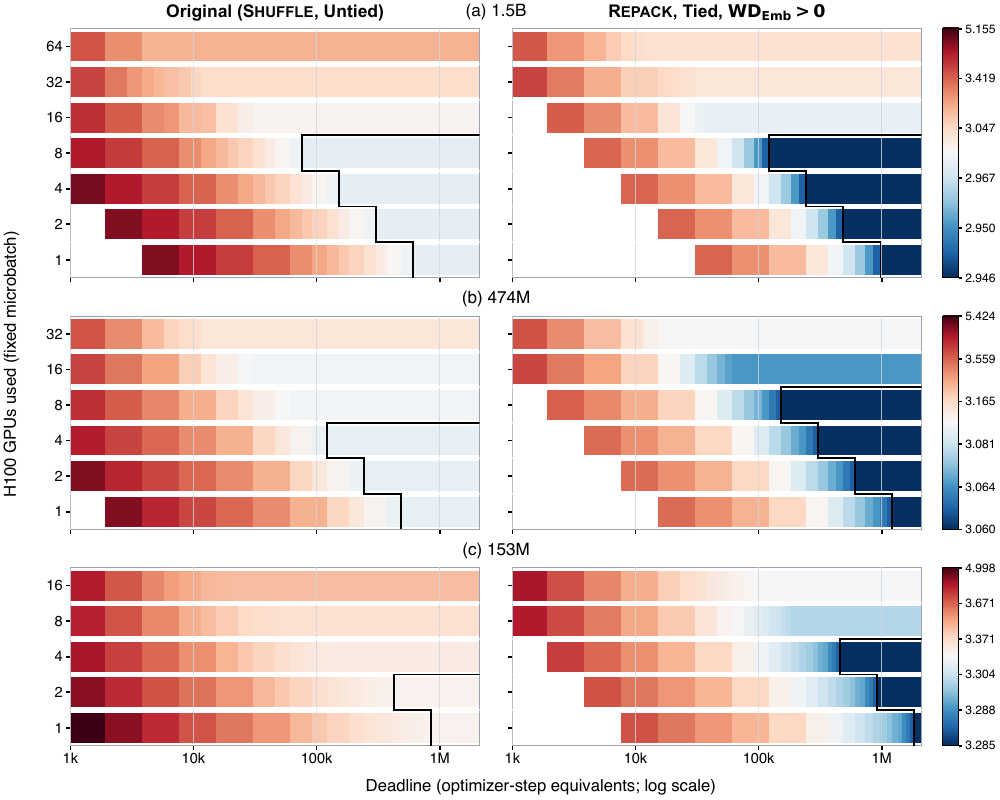}
\caption{\textbf{Joint GPU-time-constrained frontier for the original and repetition-aware recipes on the 1B-token data pool.}
Each cell shows the lowest observed validation loss achieved using exactly the indicated number of GPUs within the deadline. Colors range from blue (lower loss) to red (higher loss), normalized separately for each model size on a log scale.
The black boundary indicates where the best observed unconstrained loss is reached; beyond it, neither additional time nor additional GPUs improve validation loss.}
\label{fig:gpu_time_frontier_loss}
\end{figure*}

\subsection{Joint GPU-Time-Constrained Scaling}\label{subsec:time_gpu_constrained}
Finally, when both resources are constrained, the number of repetitions ($E$), batch size ($B$), and compute-parallelism configuration must be selected jointly; gradient accumulation provides an additional mechanism for trading training time against GPU availability.

Figure~\ref{fig:gpu_time_frontier_loss} reports the best evaluated configuration at each GPU count and deadline on the 1B-token pool, with time measured in optimizer-step equivalents. We show the resulting frontiers for \textsc{Original} (left) and repetition-aware training (right). For both recipes, more GPUs move the selected configuration toward batches that expose greater concurrency, whereas more time favors smaller batches as they generalize better with more repetitions. 
As more time becomes available, additional GPUs provide only marginal gains, and eventually, configurations using more GPUs perform worse than those using fewer GPUs, consistent with the finding in \S\ref{subsec:time_constrained_scaling}.

Comparing the left (\textsc{Original}) and right (repetition-aware) panels of Figure~\ref{fig:gpu_time_frontier_loss}, we find that repetition-aware training uniformly improves the joint GPU-time frontier. %After optimizing over all evaluated values of $E$, $B$, and compute parallelism at every budget pair, the repetition-aware recipe achieves lower validation CE than \textsc{Original} throughout the entire evaluated grid.
Its gains therefore persist even after each recipe is independently optimized for the available resources.

Finally, the solid black boundary indicates where the best observed unconstrained loss is reached. Beyond it, neither more GPUs nor more time improves the frontier.

\section{Conclusion and Limitations}
\label{sec:conclusion}
This work studies pretraining when compute is abundant but unique training data are limited, requiring models to train repeatedly on the same data until validation performance approaches its attainable minimum. We begin with the infinite-compute setting studied in prior work~\citep{kim2025infinite,xu2026dataconstrained}, but show that batch size, previously underexplored, materially affects the repeated-data frontier. We further show that this frontier can be substantially improved through repetition-aware training: dynamically repacking sequences each epoch, tying input and output embeddings, and applying positive weight decay to the shared embeddings.

However, the infinite-compute optimum can require prohibitively many sequential optimizer steps, making it impractical under tighter training deadlines. This motivates decomposing compute into two distinct resources---GPUs and time---and studying regimes in which either or both are binding. Combining observed results across 153M--1.5B-parameter models and 0.333B--3B-token data pools, we map the attainable frontier across GPU and time budgets. Our results show that optimal data-constrained pretraining requires jointly selecting the number of repetitions, batch size, compute-parallelism configuration, and training recipe.

\paragraph{Limitations.}
There are many ways to improve this work but here we highlight the two most important limitations. First, our observations are limited to 153M--1.5B-parameter models and 0.333B--3B-token data pools. In particular, the illustrative 70B/10T example requires significant extrapolation and should be viewed as illustrative projections rather than estimates that have been rigorously validated at the target scale. Second, our findings may depend on the specific architecture (particularly dense models), optimizer and learning-rate schedule (AdamW with WSD), context length, and training data (the relatively heavily filtered DCLM-Baseline corpus). Extending these results from dense models to mixture-of-experts (MoE) is especially likely to affect our conclusions and remains one of the most important directions for future data-constrained scaling research.

\section*{Acknowledgement}
We thank Will Merrill, Yashas Samaga, Mike Lewis, Iz Beltagy, Hanlin Zhang, Amil Dravid, and members of Berkeley NLP and Ai2 for discussions and feedback.

%\section*{AI Use Statement}
%Generative AI tools were used for coding and debugging, assisting with some experiments, and drafting or refining selected parts of the paper. All AI-assisted outputs were reviewed and verified by the authors. The authors take full responsibility for the methodology, results, and findings.

\section*{Reproducibility Statement}
Code and exact commands will be released upon acceptance. We report the hyperparameters (learning rate, weight decay, batch size) needed to reproduce every figure and table in \S\ref{app:experiment_configurations}.

\bibliography{iclr2027_conference}
\bibliographystyle{iclr2027_conference}

\newpage
\appendix
\section{Training, Model, and Evaluation Details}
\label{app:training_model_details}

\subsection{Model Architectures}
We train dense decoder-only Transformers with 153M, 474M, and 1.5B parameters using the \href{https://github.com/allenai/EMO}{\texttt{EMO}} codebase~\citep{wang2026emo}, built on \href{https://github.com/allenai/OLMo-core}{olmo-core}~\citep{groeneveld2024olmo}. 
All models use a 4,096-token context, a 100,278-token padded vocabulary, rotary position embeddings with base 500,000, RMSNorm, and SwiGLU feed-forward blocks. We do not use QK normalization. The attention head dimension is 64 for the 153M and 474M models and 128 for the 1.5B model.

Table~\ref{tab:model_architectures} reports the exact parameter counts and other architectural hyperparameters. After weight tying, which we find beneficial in \S\ref{sec:repetition_aware}, the parameter counts decrease to approximately 102M, 372M, and 1.3B. For consistency and to avoid confusion, we continue to refer to these models as 153M, 474M, and 1.5B throughout the paper. This convention does not unfairly favor our conclusions: it effectively compares a larger untied baseline against our smaller tied model, which nevertheless performs better. Both our experiments and prior work consistently show that larger models perform better even in the overfitted, repeated-data regime.

\begin{table*}[h]
    \caption{Model architectures. Untied counts define the nominal model labels; tied counts are the trainable parameter counts used as $N$ in our repetition-aware scaling laws. $d_{\mathrm{model}}$ is the residual width and $d_{\mathrm{ff}}$ is the SwiGLU feed-forward width.}
    \label{tab:model_architectures}
    \centering
    \small
    \begin{tabular}{@{}lrrrrrr@{}}
        \toprule
        Model & $N_{\mathrm{untied}}$ & $N_{\mathrm{tied}}$ & Layers & $d_{\mathrm{model}}$ & Heads & $d_{\mathrm{ff}}$ \\
        \midrule
        153M & 153,104,896 & 101,724,672 & 12 & 512 & 8 & 2,048 \\
        474M & 473,990,144 & 371,229,696 & 16 & 1,024 & 16 & 4,096 \\
        1.5B & 1,484,851,200 & 1,279,330,304 & 16 & 2,048 & 16 & 8,192 \\
        \bottomrule
    \end{tabular}
\end{table*}

\subsection{Data and Training Details}

All training text comes from DCLM-Baseline~\citep{li2024datacomplm}. We split documents before packing to form the training and validation sets. The validation set contains 100M tokens.

In \S\ref{sec:batch_size_two_dimensional}, we reshuffle sequence indices every epoch (defined as \ss\ in \S\ref{subsec:recipe_interventions}), by default. Later in \S\ref{subsec:recipe_interventions} we ablate this choice and converge on repacking every epoch (defined as \ds), and the rest of the paper from there uses \ds\ by default. We use seed for the dataloader (12,536 as an offset and adds epoch number) for reproducibility.

We optimize with AdamW~\citep{loshchilov2019adamw}, using $(\beta_1,\beta_2)=(0.9,0.95)$, gradient clipping at 1.0, and a $10^{-5}$ z-loss.

We use a warmup--stable--decay (WSD) learning-rate schedule~\citep{hu2024minicpm,wen2024wsd}, with an approximately 100M-token warmup and a target-specific terminal decay over the final 10\% of each run. 

We sweep batch sizes of $b\in\{16,32,64,128,256,512\}$ sequences for the 1.5B model and $b\in\{32,64,128,256,512\}$ for the 153M and 474M models. We search over peak learning rates in $\{5\times10^{-4},10^{-3},2\times10^{-3},4\times10^{-3}\}$ and weight decays in $\{0.011,0.033,0.1,0.3,1.0,3.0\}$; values outside these ranges were never selected. Because AdamW applies decay once per optimizer step, its cumulative effect depends on $B$, so we tune learning rate and weight decay jointly with batch size.
We use adaptive coordinate search: at each new epoch, the search starts from the previous best configuration and evaluates neighboring learning rates and higher weight decays.
We tune the learning rate on the 1B-token pool and transfer it to the 0.333B- and 3B-token pools, following prior evidence that the optimal learning rate is relatively stable across data-pool sizes~\citep{kim2025infinite,xu2026dataconstrained}; our preliminary experiments show the same pattern. 

\subsection{Evaluation Details}
\label{app:evaluation_details}
Our primary metric is next-token cross-entropy (CE) on a fixed, document-disjoint DCLM-Baseline validation split. \S\ref{subsec:downstream_generalization} later discusses efforts on evaluating on downstream tasks instead.

\newpage
\section{Additional Results for \S\ref{sec:batch_size_two_dimensional}}
\label{app:additional_infinite_time_results}

\subsection{Robustness to Random Seeds}
\begin{table}[h]
\centering
\small
\setlength{\tabcolsep}{5pt}
\begin{tabular}{rccccccc}
\toprule
Batch size & LR & WD & Eval.\ epoch
& Pool A & Pool B & Pool C & Across-pool SD \\
\midrule
32  & $5{\times}10^{-4}$ & 0.3   & 20
    & 3.00052 & 3.00076 & 3.00093 & 0.00021 \\
64  & $1{\times}10^{-3}$ & 0.3   & 20
    & \textbf{2.97713} & \textbf{2.97733} & \textbf{2.97594}
    & 0.00075 \\
128 & $1{\times}10^{-3}$ & 0.3   & 16
    & 2.99839 & 2.99936 & 2.99628 & 0.00157 \\
256 & $2{\times}10^{-3}$ & 0.3 & 12
    & 3.03689 & 3.04292 & 3.04167 & 0.00318 \\
\bottomrule
\end{tabular}
\caption{Impact of different random seeds in the 1B-token unique data pools. \emph{SD} indicates standard deviation.}\label{tab:independent-pool-variance}
\end{table}

We repeat experiments in \S\ref{sec:batch_size_two_dimensional} for a 1.5B model trained on a 1B-token unique-data pool, using three indepently sampled pools, denoted Pool A, Pool B, and Pool C in Table~\ref{tab:independent-pool-variance}.
Across all batches, the standard deviation of validation CE across pools ranges from only 0.00021 to 0.00318.
This variation is substantially smaller than the differences between batch configurations, and the optimal batch size is consistent across all three pools (batch size of $64$). 

\subsection{Comparison with Unique-Data Training}
\begin{figure}[h]
    \centering
    \includegraphics[width=0.6\linewidth]{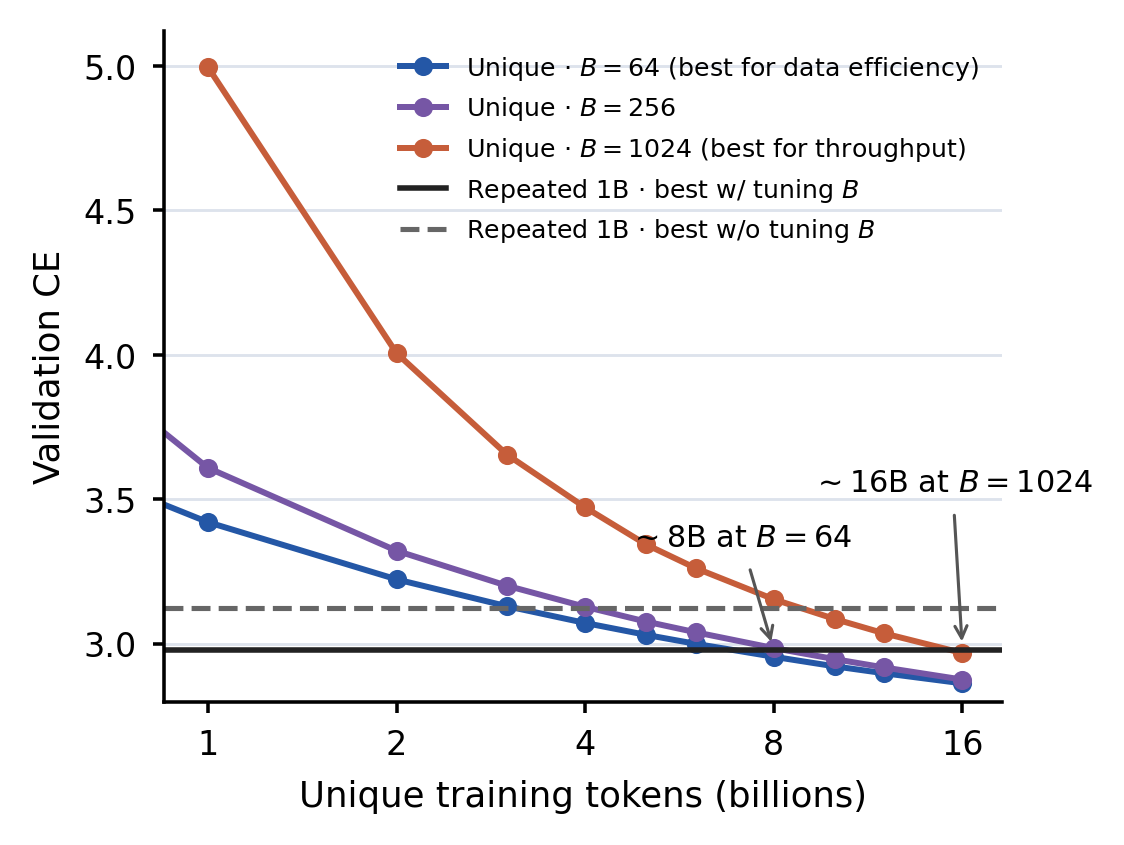}
    \caption{\textbf{Unique-data equivalent of repeated-data training.} Curves show validation CE for unique-data training at three batch sizes. Horizontal lines mark the best repeated 1B-token results with and without tuning $B$. With tuning, repeated-data training matches $6\times$--$8\times$ unique data at $B=64$ and $\sim16\times$ at $B=1024$; without tuning, it matches $\sim8\times$ at $B=1024$.
    }\vspace{-1em}
    \label{fig:unique_equivalence_1b_pool}
\end{figure}
\oldparagraph{Repeated-data training substitutes for substantially more unique data.}
Figure~\ref{fig:unique_equivalence_1b_pool} compares the best repeated-data endpoints with batch-tuned unique-data curves. At matched $B=64$, the batch-tuned repeated-data recipe matches training on $\sim8\times$ more unique data. Against the throughput-oriented $B=1024$ unique-data curve, it matches $\sim16\times$ more unique data. Without tuning $B$, the prior repeated-data recipe matches only $\sim8\times$ more unique data. Tuning $B$ therefore provides a gain comparable to doubling the unique-data budget.

Prior findings tied to a particular number of epochs are also batch-dependent. For example, the observation that unique and repeated data remain similar for roughly four epochs~\citep{muennighoff2023scaling} changes with $B$: larger batches keep the two settings close for more epochs but fewer optimizer steps, while smaller batches keep them close for more optimizer steps but fewer epochs.

\subsection{Downstream Task Generalization}
\label{subsec:downstream_generalization}

\begin{figure*}[t]
    \centering
    \includegraphics[width=0.9\textwidth]{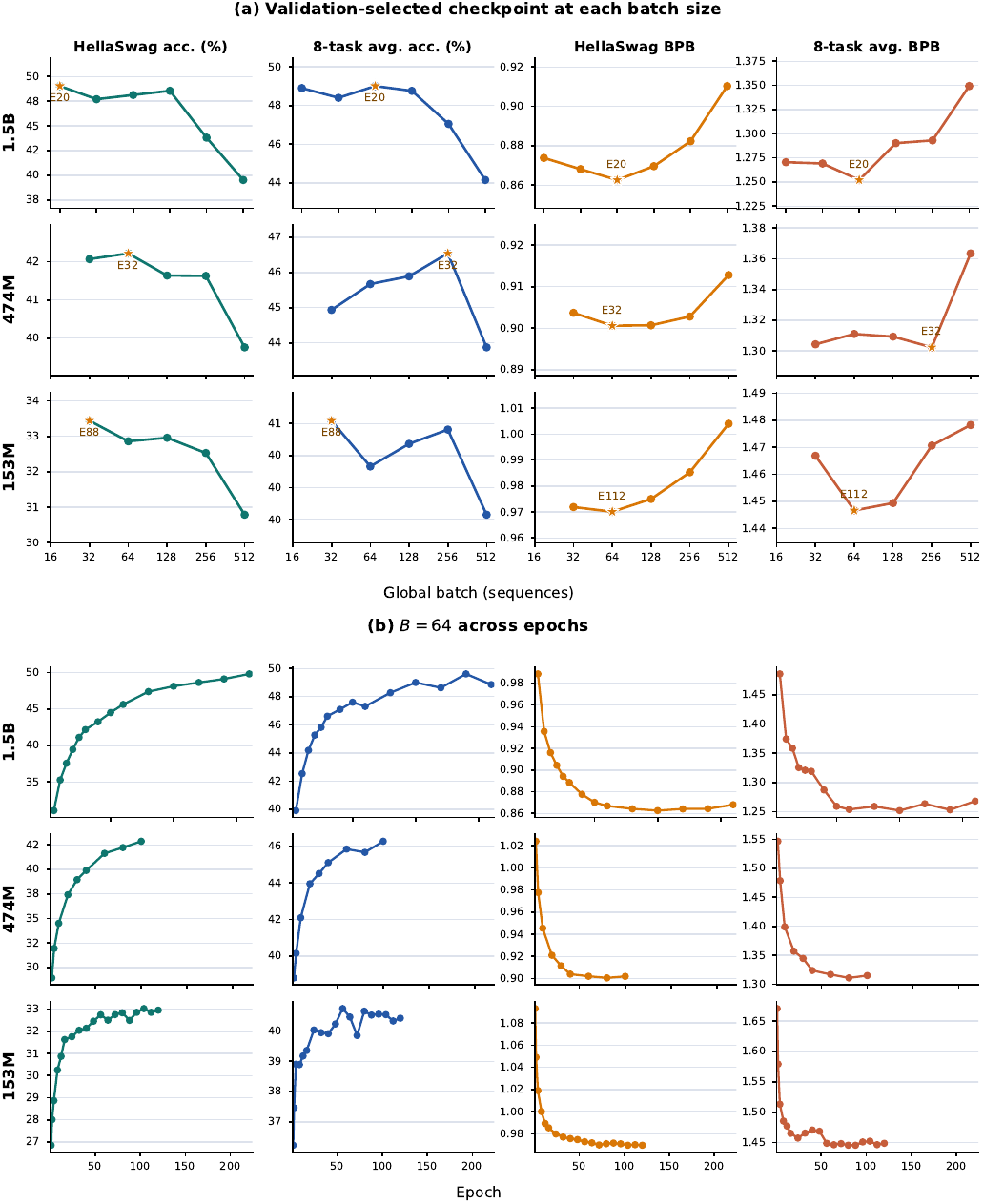}
    \caption{\textbf{HellaSwag and aggregate downstream performance.} (a) Performance versus batch size. Each best available epoch was selected based on validation CE. (b) Performance across epochs at $B=64$.}
    \label{fig:downstream_combined}
\end{figure*}

Making modeling conclusions from downstream performance is substantially more difficult than doing so with validation loss: individual-task accuracies vary considerably across checkpoints, particularly for small models. Recent work therefore predicts task performance indirectly through language-modeling loss, aggregates results across tasks, or uses soft metrics like perplexity or bits per byte (BPB)~\citep{gadre2024language,bhagia2025taskscaling,heineman2025signal}. These studies find that task aggregates are more predictable than many individual accuracies and that perplexity and BPB are often more reliable.

We evaluate eight widely used pre-training evaluation benchmarks: HellaSwag~\citep{zellers2019hellaswag}; ARC-Easy and ARC-Challenge~\citep{clark2018arc}; CommonsenseQA~\citep{talmor2019commonsenseqa}; OpenBookQA~\citep{mihaylov2018openbookqa}; PIQA~\citep{bisk2020piqa}; SocialIQA~\citep{sap2019socialiqa}; and WinoGrande~\citep{sakaguchi2020winogrande}.

Figure~\ref{fig:downstream_combined} reports performance of three models, 1.5B, 474M, and 153M, on four downstream task metrics: Hellaswag accuracy, aggregated eight-task averaged accuracy, Hellaswag BPB, and aggregated eight-task BPB. The top figure (a) reports the metric from checkpoints selected by validation CE for each batch size, and the bottom figure (b) reports the metric over epochs for $B=64$.

The main conclusions from the earlier sections continue to hold. Across all model sizes, overly large batches perform poorly; very small batches are not uniformly optimal either; the strongest results tend to occur at intermediate batch sizes.

At the same time, we confirm previous findings that downstream evaluation exhibits substantially greater variance. Its curves are not always convex, as would be expected from a reliable scaling signal. BPB generally produces smoother trends than accuracy, consistent with prior work~\citep{heineman2025signal}, e.g., convex trends emerge in both BPB panels for the 1.5B and 153M models and in the HellaSwag BPB panel for the 474M model. Whenever the curve is convex, the downstream results closely agree with the validation-loss results: the optimal batch size lies consistently near $B=64$ across model sizes.

\newpage
\section{Additional Results for \S\ref{sec:repetition_aware}}\label{app:repetition_aware}

\myskip{
\begin{figure*}[t]
\centering
\includegraphics[width=\textwidth]{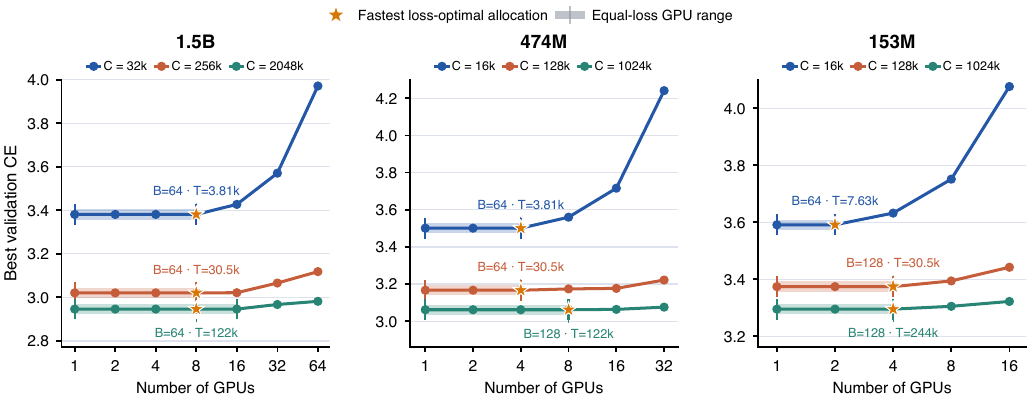}
\caption{\textbf{Optimal GPU and time allocation under a fixed total compute budget for the 1B-token pool.} Time $T$ is measured in model-specific optimizer-step equivalents. Highlighted ranges achieve the same minimum loss; stars indicate the fastest allocation, with ties broken in favor of fewer GPUs. Smaller batch sizes can be preferred under limited compute, while larger batch sizes become preferable as the budget increases, but capped to the infinite-compute optimal batch size identified in \S\ref{subsec:recipe_interventions_results} and Figure~\ref{fig:recipe_batch_optima} ($B=64$ for the 1.5B model and $B=128$ for the 474M and 153M models).}
\label{fig:fixed_compute_gpu_allocation}
\end{figure*}

\paragraph{Optimal resource allocation under fixed compute}
What if we relax the infinite compute assumption, and consider a fixed total compute budget, measured as GPU count $\times$ training time, which can be freely allocated between the two?
Figure~\ref{fig:fixed_compute_gpu_allocation} shows that smaller budgets typically favor fewer GPUs, allowing longer training and more optimizer steps. As the budget increases the optimal batch size grows, but capped to the infinite-compute optimal batch size identified earlier. %in \S\ref{subsec:recipe_interventions_results} and Figure~\ref{fig:recipe_batch_optima} ($B=64$ for the 1.5B model and $B=128$ for the 474M and 153M models).

\paragraph{Comparison with existing open-source defaults.}
}

\subsection{Comparison with existing open-source defaults}\label{app:comparison-with-open-source-defaults}
Our survey of nine representative open-source training stacks (Table~\ref{tab:open_source_defaults}) shows that their default configurations vary substantially and often differ from our best-performing repetition-aware training. We recommend that researchers and LLM developers treat these choices as explicit design decisions rather than uncritically inheriting a stack's defaults.

\begin{center}
\begin{minipage}{\linewidth}
    \centering 
    \vspace{-.5em}
    \footnotesize    \setlength{\tabcolsep}{3pt}    \renewcommand{\arraystretch}{1.08}    
\begin{tabularx}{.8\linewidth}{@{}>{\raggedright\arraybackslash}p{0.24\linewidth}>{\raggedright\arraybackslash}Xccc@{}}        \toprule        Stack & Repeated-data order & Tying & Embed. WD & Head WD \\        \midrule        

\href{https://github.com/mlfoundations/open_lm/blob/9bb92ef1689333534b7057942a20d18a46d1fa52/open_lm/data.py\#L340-L410}{OpenLM} & \fixed & No & $>0$ & $>0$ \\       
\href{https://github.com/marin-community/marin/blob/deee9eabb335772460de9237e17362ce094c127c/lib/levanter/src/levanter/data/text/datasets.py\#L388-L405}{Marin / Levanter} & \fixed & No & 0 & $>0$ \\

\href{https://github.com/allenai/OLMo-core/blob/b7e9671d7ea48af94838c4f124703c3ae36f0c70/src/olmo_core/data/data_loader.py\#L635-L684}{olmo-core} & \ss & No & 0 & $>0$ \\     
\href{https://github.com/mosaicml/llm-foundry}{LLM Foundry MPT} & \ss & Yes & \multicolumn{2}{c}{shared 0} \\        \href{https://github.com/huggingface/transformers}{Hugging Face Trainer} & \ss & Model-specific & 0 & 0 \\   
\href{https://github.com/facebookresearch/lingua/blob/437d680e521873bb5971067148a69587790da853/lingua/data.py\#L319-L392}{Meta Lingua} & \ss & No & $>0$ & $>0$ \\        %\href{https://github.com/pytorch/torchtitan/blob/0d2438f954c2581cbf22b45f45835bbfa8e0c8db/torchtitan/hf_datasets/text_datasets.py\#L129-L186}{TorchTitan} & \ds & Yes & \multicolumn{2}{c}{shared $>0$} \\    

\href{https://github.com/EleutherAI/pythia/blob/a19eecb807ec2c79a39ebf18108816e6ffffc1d5/models/160M/pythia-160m.yml\#L8-L15}{Pythia / GPT-NeoX} & \ds & No & $>0$ & $>0$ \\  
\href{https://github.com/NVIDIA/Megatron-LM/blob/c374e5b65e32b498763e3666b40490314b554cb7/megatron/core/datasets/gpt_dataset.py\#L500-L601}{Megatron-LM} & \ds & Yes & \multicolumn{2}{c}{shared $>0$} \\        \href{https://github.com/NVIDIA-NeMo/Megatron-Bridge/blob/6d9a87d6a0ba7550fdd9455ef2a97a62a7dcdef3/src/megatron/bridge/models/gpt_provider.py\#L128-L142}{NeMo Megatron Bridge} & \ds & Model-specific & $>0$ & $>0$ or shared \\           

\midrule        \textbf{Ours: data-constrained} & \textbf{\ds} & \textbf{Yes} & \multicolumn{2}{c}{\textbf{shared $>0$}} \\        

\bottomrule    \end{tabularx}  
    \vspace{-.5em}
    \captionsetup{hypcap=false}\captionof{table}{Representative open-source defaults and our data-constrained recommendation.}    \label{tab:open_source_defaults}
    % \vspace{-.5em}
\end{minipage}
\end{center}\FloatBarrier

\subsection{Failed Attempts or Higher-risk Recipes: Small-Batch Noise at Large Batch}\label{app:higher_risk_improvements}

One of our key findings is that the gradient noises from small-batch training likely provide beneficial regularization. This raises a natural question: can we recover some of these regularization benefits while retaining the computational efficiency and parallelism of larger-batch training? In other words, can we delay overfitting without sacrificing the parallelism for faster training?

We explored several approaches in this direction.

\paragraph{Batch warmdown.} This method gradually reduces the batch size over the course of training. However, in our experiments, it did not improve the training-time-validation-loss Pareto frontier.

\paragraph{Branch-train-merge.} Following \citet{li2022branchtrainmerge}, we train multiple models initialized from a shared checkpoint and subsequently merges their weights. However, in our experiments, it did not improve the training-time-validation-loss Pareto frontier.

\paragraph{Structured gradient noise.} We add structured gradient noise to large-batch updates. However, it suffered from recurrent gradient spikes and did not yield a stable training trajectory.

\paragraph{Periodically synchronized local optimization.}
Among the approches we tested, this was the only one that produced promising results. The general idea is to perform multiple local optimization steps in parallel, and then synchronizing across workers.
We consider two concrete methods: (1) LocalSGD~\citep{stich2019local}, which allows each worker to perform $H$ local optimization steps and then averaging model parameters across workers, and (2) DiLoCo~\citep{douillard2023diloco}, which similarly performs multiple local updates between synchronization points, but uses an outer optimizer to update the shared model from the accumulated differences between the initial and locally trained parameters.

In our experiments, DiLoCo consistently outperformed LocalSGD, so we report DiLoCo's results here.
We fix the synchronization interval to $H=32$, the outer learning rate to $0.7$, and the outer momentum to $0.9$. We vary the per-worker batch size over ${64,256}$ and initialize from checkpoints obtained after either two or four passes over the data using a global batch size of 1,024, yielding four configurations. After tuning the inner learning rate and weight decay, we report the best result across these configurations for each metric and matched optimizer-step budget.

\begin{table}[h]
    \centering
    \footnotesize
    \setlength{\tabcolsep}{6pt}
\begin{tabular}{r l r r r r r}
        \toprule
        & & & \multicolumn{2}{c}{HellaSwag} & \multicolumn{2}{c}{8-task average} \\
        \cmidrule(lr){4-5}\cmidrule(lr){6-7}
        Steps & Method & Val. CE $\downarrow$ & Acc. $\uparrow$ & BPB $\downarrow$
              & Acc. $\uparrow$ & BPB $\downarrow$ \\
        \midrule
        1,907 & $B=512$ \,{\scriptsize(E4)}  & 3.221          & 35.02          & 0.9374          & 42.70          & 1.4018          \\
        1,907 & $B=1{,}024$ \,{\scriptsize(E8)} & 3.226          & 36.41          & 0.9386          & 42.55          & 1.4265          \\
        1,907 & $B=1{,}024$ w. DiLoCo \,{\scriptsize(E8)}     & \textbf{3.217} & \textbf{38.09} & \textbf{0.9331} & \textbf{43.17} & \textbf{1.3831} \\
        \midrule
        2,861 & $B=512$ \,{\scriptsize(E6)}   & ---            & ---            & ---            & ---            & ---            \\
        2,861 & $B=1{,}024$ \,{\scriptsize(E12)} & \textbf{3.160} & 37.64          & 0.9191          & 43.48          & 1.3673          \\
        2,861 & $B=1{,}024$ w. DiLoCo \,{\scriptsize(E12)}    & 3.170          & \textbf{40.72} & \textbf{0.9157} & \textbf{44.62} & \textbf{1.3596} \\
        \midrule
        3,815 & $B=512$ \,{\scriptsize(E8)}   & 3.137          & 40.50          & 0.9094          & 44.93          & 1.3488          \\
        3,815 & $B=1{,}024$ \,{\scriptsize(E16)} & \textbf{3.128} & 39.86          & 0.9083          & 45.16          & 1.3614          \\
        3,815 & $B=1{,}024$ w. DiLoCo \,{\scriptsize(E16)}    & 3.139          & \textbf{42.71} & \textbf{0.9065} & \textbf{45.88} & \textbf{1.3301} \\
        \midrule
        5,722 & $B=512$ \,{\scriptsize(E12)}  & \textbf{3.091} & 40.71          & \textbf{0.8980} & 45.72          & \textbf{1.3078} \\
        5,722 & $B=1{,}024$ \,{\scriptsize(E24)} & 3.122          & 43.20          & 0.9042          & 45.56          & 1.3655          \\
        5,722 & $B=1{,}024$ w. DiLoCo \,{\scriptsize(E24)}    & 3.128          & \textbf{45.81} & 0.9029          & \textbf{46.97} & 1.3435          \\
        \bottomrule
    \end{tabular}
    \vspace{2pt}
    \caption{Matched-optimizer-step comparison of baselines and DiLoCo. All methods use \ds.
    Bold marks the best available
    result within each step budget.}
    \label{tab:diloco_matched_steps}
\end{table}

Table~\ref{tab:diloco_matched_steps} shows that DiLoCo does not consistently improve validation CE. It does, however, consistently improve downstream accuracy: on HellaSwag, our most reliable downstream metric, it gains approximately 1.7--3.1\% over the stronger matched-step baseline. This was one of few settings in which validation CE and downstream performance consistently led to different conclusions.

We therefore view DiLoCo as promising but not yet conclusive, so did not use it for the rest of the experiments in the paper. We encourage further investigation into local-update methods for repeated-data training.

\newpage
\section{Additional Results for \S\ref{sec:unified_gpu_time_scaling}}
\label{app:section5_supplementary}

\subsection{Illustrative 70B/10T Repetition and Batch Extrapolation}
\label{app:loss_fitting_scaling_law}

We use our measured grid to estimate the repetition count and batch size---and hence the time and GPU resources---for a hypothetical 70B model trained on a 10T-token unique-data pool. This is an extreme, illustrative extrapolation rather than a precise target-scale prediction.

\paragraph{Epoch count.}
We first fit the validation-loss curves only to the tuned repetition-aware trajectories---\ds, tied embeddings, and positive embedding weight decay---for the $3\times3$ grid of model sizes and unique-data pools in \S\ref{sec:repetition_aware}. The fit uses the exact tied-embedding trainable counts $N\in\{101{,}724{,}672;\ 371{,}229{,}696;\ 1{,}279{,}330{,}304\}$ from Table~\ref{tab:model_architectures} and the highest-precision stored validation CE from all 100 eligible checkpoints.

We model validation loss after $E$ passes over the unique pool as
\begin{equation}
\widehat L(E,N,\Du)
=
L_0+A E^{-\alpha}+B_{\mathrm{rep}}E^{\beta},
\qquad
B_{\mathrm{rep}}=
\frac{\alpha A}{\beta E_{\min}^{\alpha+\beta}},
\label{eq:repetition_aware_loss_law}
\end{equation}
where the first term captures useful learning and the second captures degradation from repetition; this parameterization makes $E_{\min}$ the continuous minimizer.

The loss is the mean of the weighted squared errors across the nine grid cells, in order to give each cell equal weight; within each cell, checkpoints are weighted by their local spacing in $\log E$ so that densely sampled early epochs do not dominate.
Because the loss curves become very flat near saturation, estimates of $E_{\min}$ can be brittle. We therefore add soft hinge penalties that capture the observed curve shape without introducing synthetic loss values. When the first non-improving checkpoint brackets the turn between evaluated epochs $E_a$ and $E_b$, the penalty is zero for $E_a < E_{\min} < E_b$ and grows quadratically only when $E_{\min}$ falls outside this interval.
For $N=153$M and $D_u=3$B, we stopped training at $E=640$ due to compute constraints, even though the validation loss was still improving; we therefore use a one-sided penalty that increases only when $E_{\min}$ lies below that endpoint. The hinge weight is 2.0. 

Let $x=\log(N/371{,}229{,}696)$ and $y=\log(\Du/10^9)$. A joint 15-parameter fit makes the five latent curve parameters affine in $(x,y)$:
\begin{equation}
\begin{bmatrix}
L_0\\ \log A\\ \log\alpha\\ \log E_{\min}\\ \log\beta
\end{bmatrix}
=
\begin{bmatrix}
-0.10524 & -0.59081 & -1.51710\\
 0.60213 &  0.18444 &  0.39532\\
-1.62858 & -0.04994 & -0.64827\\
 4.55678 & -0.74715 &  0.54148\\
-2.80029 &  0.25300 & -0.07690
\end{bmatrix}
\begin{bmatrix}1\\x\\y\end{bmatrix}.
\label{eq:joint_loss_parameterization}
\end{equation}
The resulting mean within-grid, log-spacing-weighted $R^2$ for the checkpoint-loss predictions is $0.980$. 

Next, we use the fitted curves to estimate $E^{\star}(N,\Du)$, the practical saturation epoch for model size $N$ and unique-data pool $\Du$. Because the exact minimizer can be unstable when the loss curve is flat, we define the stopping target as the first integer epoch within 0.001 validation loss of the fitted minimum:
\begin{equation}
E^{\star}(N,\Du)
=
\min\!\left\{E\ge1:\;\widehat L(E,N,\Du)\le \widehat L(E_{\min},N,\Du)+0.001\right\}.
\label{eq:near_optimal_epoch_definition}
\end{equation}
This corresponds to a perplexity ratio of at most $\exp(0.001)\approx1.001$.

To extrapolate beyond the measured grid, we calculate $E^{\star}$ for each of the nine fitted grid-cell curves and fit a power function to those nine model-derived values:
\begin{equation}
E^{\star}(N,\Du)
=
37.64
\left(\frac{N}{10^9}\right)^{-0.6986}
\left(\frac{\Du}{10^9}\right)^{0.5111}.
\label{eq:appendix_epoch_extrapolation}
\end{equation}
\label{eq:epoch_ratio_extrapolation}
This summary fit has a multiplicative RMS error of $1.011\times$ and log-space $R^2=0.9998$ on the nine model-derived targets, and gives $E^{\star}(70\mathrm{B},10\mathrm{T})=214.2$, which we round to 214 epochs.

\paragraph{Batch size.}
For batch size, nine observations are too few to support even this simple fit reliably. We therefore add the three Pool-111M batch optima, producing twelve observations for the batch-size fit only. The Pool-111M runs were continued only far enough to identify a clear batch-size winner for each model; because we did not train every model to saturation on that pool, we exclude those trajectories from the loss and epoch fits above. Ordinary least squares in log space gives
\begin{equation}
B^{\star}(N,\Du)
=
101.751
\left(\frac{N}{10^9}\right)^{-0.13583}
\left(\frac{\Du}{10^9}\right)^{0.35740}
\quad\text{sequences}.
\label{eq:appendix_batch_extrapolation}
\end{equation}
\label{eq:recipe_batch_extrapolation}
Its multiplicative RMS error is $1.29\times$ and its log-space $R^2$ is $0.768$. For a 70B model and 10T unique tokens, this fit extrapolates to approximately $1{,}536$ sequences. %, or $6.29$ million tokens per optimizer update at a 4,096-token context. %The batch-size extrapolation remains the larger source of uncertainty in the large-scale resource estimates.

\paragraph{Resulting resource estimate.}
Rounding the projections to $E^{\star}=214$ and $B^{\star}=1{,}536$ gives approximately 340.1 million optimizer steps. Under the conventional configuration (Table~\ref{tab:throughput-vs-latency-parallelism})---gradient accumulation $a=1$, microbatch size $m=1$, tensor parallelism $t=4$, pipeline parallelism $p=1$, and optimizer-step latency $\tau(70)\approx2.10$ seconds---this corresponds to approximately 6,144 H100 GPUs and 22.6 years of training.

\subsection{From Throughput-First to Latency-First Compute Parallelism}\label{subsec:aggressive_model_parallelism}

This section describes how explicitly replacing a conventional, throughput-optimal model-parallelism configuration with a latency-optimal configuration can reduce training time---at the cost of using more GPUs---when training is time-constrained.

\paragraph{Conventional throughput-first policy.}
Large-scale training distributes a global batch across a compute-parallelism configuration defined by the local microbatch size and the data-parallel (DP), tensor-parallel (TP), and pipeline-parallel (PP) degrees.\footnote{Since our context length is fixed at 4,096 tokens, we leave context parallelism outside the scope.}
With gradient accumulation fixed to $a=1$, so that $B_{\rm fixed}=dm$, the conventional policy selects the feasible configuration with the highest throughput per GPU:
\[
(d_{\rm th}^*,t_{\rm th}^*,p_{\rm th}^*,m_{\rm th}^*)=
\underset{\substack{d,t,p,m\\dm=B_{\rm fixed}\\dtp\leq G}}{\arg\max}\;
\frac{mS}{tp\,T_{\rm step}(d,t,p,m)}.
\]
In practice, this favors the minimum model-parallel degree needed for memory feasibility and satisfactory utilization, a microbatch near the per-GPU throughput optimum, and DP for the remaining batch parallelism.

\paragraph{Latency-first policy.}
The latency-first policy instead selects the feasible microbatch and DP/TP/PP configuration with the shortest optimizer step while preserving the same global batch:
\[
(d^*,t^*,p^*,m^*)=
\underset{\substack{d,t,p,m\\dm=B_{\rm fixed}\\dtp\leq G}}{\arg\min}\;
T_{\rm step}(d,t,p,m),
\]
subject to memory, divisibility, and topology constraints. Additional H100s are assigned to TP or PP only when the reduction in exposed DP communication exceeds the added model-parallel communication and scheduling overhead.

\begin{table*}[t]
\centering
\scriptsize
\setlength{\tabcolsep}{4pt}
\begin{tabular}{lrrrrrrr@{\qquad}rrrrrrr}
\toprule
&\multicolumn{7}{c}{Conventional throughput-first H100}
&\multicolumn{7}{c}{Latency-first H100}\\
\cmidrule(lr){2-8}\cmidrule(lr){9-15}
Model &$m$&DP&TP&PP&GPUs&$T_{\rm step}$&Tok./s/GPU
      &$m$&DP&TP&PP&GPUs&$T_{\rm step}$&Tok./s/GPU\\
\midrule
1.5B&8&128&1&1&128 &$\sim0.60$ s&$54.6$k
    &1&1024&1&1&1024 &$\sim0.22$ s&$18.6$k\\
8B  &2&512&1&1&512 &$\sim1.00$ s&$8.19$k
    &1&1024&2&1&2048 &$\sim0.60$ s&$3.41$k\\
32B &1&1024&2&1&2048 &$\sim1.95$ s&$1.05$k
    &1&1024&4&2&8192 &$\sim0.70$ s&$0.73$k\\
70B &1&1024&4&1&4096 &$\sim2.10$ s&$0.49$k
    &1&1024&8&4&32768 &$\sim0.56$ s&$0.23$k\\
\bottomrule
\end{tabular}
\caption{H100 projections for throughput-first and latency-first compute-parallelism configurations at the same global batch size, $B=1{,}024$ sequences, with gradient accumulation $a=1$. Throughput and latency are simulated using a model that accounts for communication overhead. See \S\ref{app:parallelism-estimation} for details.}
\label{tab:throughput-vs-latency-parallelism}
\end{table*}

Implementing and benchmarking the full microbatch and DP/TP/PP design space is beyond our scope. Instead, we use the simulated, communication-aware H100 model in \S\ref{app:parallelism-estimation} to estimate optimizer-step latency ($T_{\rm step}$) and throughput (tokens/s/GPU) across feasible configurations, and then report the throughput-optimal (left) and latency-optimal (right) configurations in Table~\ref{tab:throughput-vs-latency-parallelism}. These results are fitted projections rather than end-to-end measurements.

Results show that explicitly optimizing for latency yields $1.67\times$--$3.75\times$ shorter optimizer steps across model sizes, at the cost of using more GPUs, through smaller microbatches and more aggressive model parallelism. For example, the latency-optimal 70B configuration reduces step latency by $3.75\times$ by using $8\times$ more GPUs. It is also worth noting that further model parallelism eventually stops improving latency because of communication and other overheads. For example, the 70B latency improvement saturates at $3.75\times$ under our fitted model; beyond that point, additional model parallelism does not reduce latency.

\subsection{Supplementary Materials for Time--GPU-Constrained Scaling}
\label{app:time_gpu_constrained_supplementary}
\begin{figure*}[h]
    \centering
    \includegraphics[width=\textwidth]{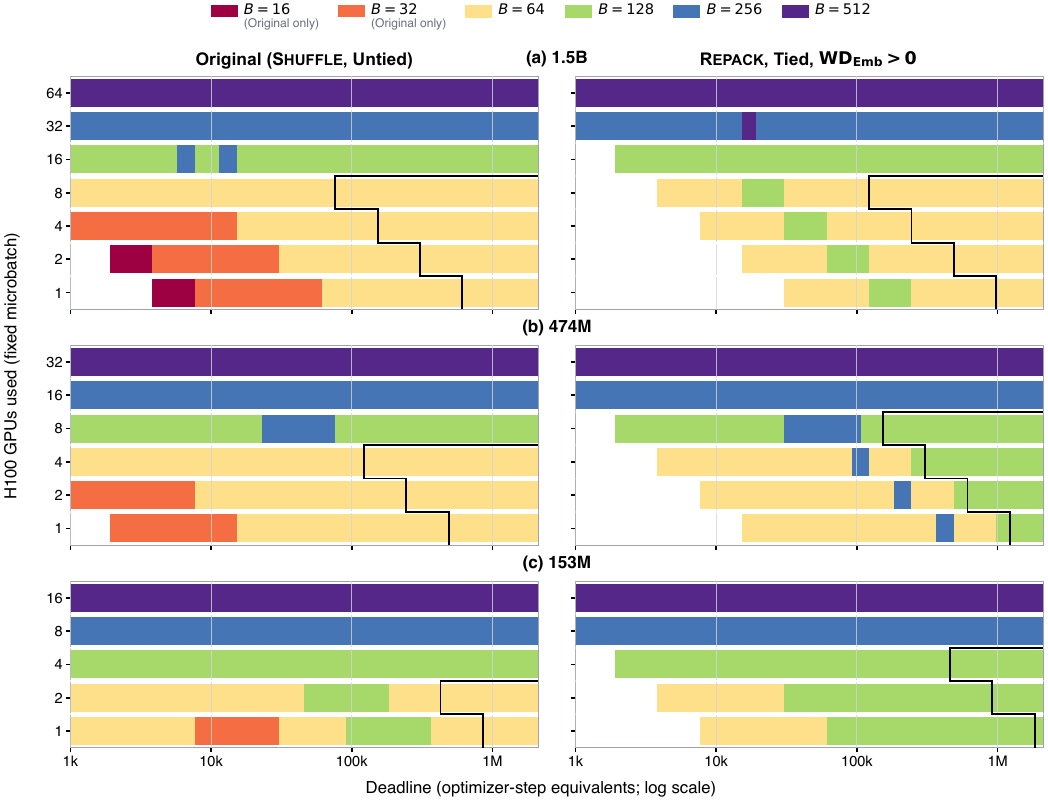}
    \caption{\textbf{Optimal batch size at each GPU count and deadline on the 1B-token pool, both for the original (left) and the repetition-aware recipe (right).}
    Colors indicate the batch size of the observed configuration with the lowest validation CE that uses exactly the indicated number of H100 GPUs and finishes within the optimizer-step-equivalent deadline, with microbatch size held fixed. 
    The black boundary indicates the region where the globally optimal loss under unconstrained GPU and time budgets is reached; beyond it, moving right (more time) or up (more GPUs) does not improve the validation loss.
    }
    \label{fig:gpu_time_frontier_batch}
\end{figure*}

Figure~\ref{fig:gpu_time_frontier_batch} shows the batch size selected at each GPU count and deadline for the loss frontier in Figure~\ref{fig:gpu_time_frontier_loss} (\S\ref{subsec:time_gpu_constrained}). Each configuration uses exactly the indicated number of GPUs, with microbatch size held fixed. 
The corresponding validation losses are shown in Figure~\ref{fig:gpu_time_frontier_loss}, with the black boundary indicating the region where the globally optimal loss under unconstrained GPU and time budgets is reached; beyond it, moving right (more time) or up (more GPUs) does not improve the validation loss.

%\newpage
%\input{sections/d_time_constrained_improvements}

\newpage

\section{Experiment Configurations for Sections~\ref{sec:batch_size_two_dimensional} and~\ref{sec:repetition_aware}}
\label{app:experiment_configurations}

We report the exact hyperparameters used in all experiments in the main paper. $B$ is the global batch size in 4,096-token sequences, LR is the peak learning rate, WD is AdamW weight decay, and $E$ is the number of training passes, including terminal WSD decay.
% WD$_{\rm Emb}$ denotes decay on the input embedding or, when tied, the shared input/output matrix.
\begin{table}[!htbp]
\centering\myfontsize
\captionsetup{font=footnotesize,skip=4pt}
\setlength{\tabcolsep}{2pt}
\renewcommand{\arraystretch}{1.02}
\caption{Hyperparameters for the batch-size sweep in Figure~\ref{fig:batch_optima_1b_pool} (\S\ref{sec:batch_size_two_dimensional}), all on a 1B-token pool. The 1.5B, $B=64$ row also supplies our observed result in the batch-size comparison in Table~\ref{tab:compute_optimal_batch_comparison}. All runs use the original recipe.}
\label{tab:section3_selected_configs}
\begin{minipage}[t]{.325\linewidth}
\centering\myfontsize
\textbf{1.5B}\par\smallskip
\begin{tabular}{@{}rrrrr@{}}
\toprule
$B$ & LR & WD & $E$ & CE \\
\midrule
16 & $5\!\times\!10^{-4}$ & 0.1 & 20 & 3.017 \\
32 & $5\!\times\!10^{-4}$ & 0.3 & 20 & 3.001 \\
64 & $10^{-3}$ & 0.3 & 20 & 2.977 \\
128 & $10^{-3}$ & 0.3 & 16 & 2.998 \\
256 & $2\!\times\!10^{-3}$ & 0.3 & 12 & 3.037 \\
512 & $2\!\times\!10^{-3}$ & 0.3 & 8 & 3.124 \\
\bottomrule
\end{tabular}
\end{minipage}\hfill
\begin{minipage}[t]{.325\linewidth}
\centering\myfontsize
\textbf{474M}\par\smallskip
\begin{tabular}{@{}rrrrr@{}}
\toprule
$B$ & LR & WD & $E$ & CE \\
\midrule
32 & $10^{-3}$ & 0.1 & 32 & 3.113 \\
64 & $2\!\times\!10^{-3}$ & 0.1 & 32 & 3.093 \\
128 & $2\!\times\!10^{-3}$ & 0.3 & 40 & 3.101 \\
256 & $2\!\times\!10^{-3}$ & 0.3 & 32 & 3.102 \\
512 & $2\!\times\!10^{-3}$ & 0.3 & 24 & 3.136 \\
\bottomrule
\end{tabular}
\end{minipage}\hfill
\begin{minipage}[t]{.325\linewidth}
\centering\myfontsize
\textbf{153M}\par\smallskip
\begin{tabular}{@{}rrrrr@{}}
\toprule
$B$ & LR & WD & $E$ & CE \\
\midrule
32 & $10^{-3}$ & 0.1 & 88 & 3.350 \\
64 & $2\!\times\!10^{-3}$ & 0.1 & 112 & 3.332 \\
128 & $2\!\times\!10^{-3}$ & 0.1 & 72 & 3.341 \\
256 & $2\!\times\!10^{-3}$ & 0.1 & 48 & 3.360 \\
512 & $2\!\times\!10^{-3}$ & 0.1 & 40 & 3.416 \\
\bottomrule
\end{tabular}
\end{minipage}
\end{table}

\begin{table}[!htbp]
\centering\myfontsize
\captionsetup{font=footnotesize,skip=4pt}
\setlength{\tabcolsep}{2pt}
\renewcommand{\arraystretch}{1.02}
\caption{Examples of selected hyperparameters across training horizons for three model sizes on the 1B-token pool, corresponding to Figure~\ref{fig:batch_optima_1b_pool} (\S\ref{sec:batch_size_two_dimensional}). All evaluated epochs through the first checkpoint after the validation loss saturates or reverses are shown. Bold marks the best checkpoint. All runs use the original recipe.}
\label{tab:section3_epoch_configurations}
\begin{minipage}[t]{.325\linewidth}
\centering\myfontsize
\textbf{1.5B, $B=512$}\par\smallskip
\begin{tabular}{@{}rrrr@{}}
\toprule
$E$ & LR & WD & CE \\
\midrule
1 & $10^{-3}$ & 0.1 & 4.000 \\
2 & $10^{-3}$ & 0.3 & 3.469 \\
4 & $10^{-3}$ & 0.3 & 3.224 \\
\textbf{8} & $\boldsymbol{2\!\times\!10^{-3}}$ & \textbf{0.3} & \textbf{3.124} \\
12 & $10^{-3}$ & 0.3 & 3.198 \\
\bottomrule
\end{tabular}
\end{minipage}\hfill
\begin{minipage}[t]{.325\linewidth}
\centering\myfontsize
\textbf{474M, $B=512$}\par\smallskip
\begin{tabular}{@{}rrrr@{}}
\toprule
$E$ & LR & WD & CE \\
\midrule
1 & $2\!\times\!10^{-3}$ & 0.1 & 4.235 \\
2 & $2\!\times\!10^{-3}$ & 0.1 & 3.625 \\
4 & $2\!\times\!10^{-3}$ & 0.3 & 3.364 \\
8 & $2\!\times\!10^{-3}$ & 0.3 & 3.226 \\
12 & $2\!\times\!10^{-3}$ & 0.3 & 3.178 \\
16 & $2\!\times\!10^{-3}$ & 0.3 & 3.155 \\
\textbf{24} & $\boldsymbol{2\!\times\!10^{-3}}$ & \textbf{0.3} & \textbf{3.136} \\
32 & $2\!\times\!10^{-3}$ & 0.3 & 3.136 \\
\bottomrule
\end{tabular}
\end{minipage}\hfill
\begin{minipage}[t]{.325\linewidth}
\centering\myfontsize
\textbf{153M, $B=256$}\par\smallskip
\begin{tabular}{@{}rrrr@{}}
\toprule
$E$ & LR & WD & CE \\
\midrule
1 & $2\!\times\!10^{-3}$ & 0.033 & 4.038 \\
2 & $2\!\times\!10^{-3}$ & 0.033 & 3.710 \\
4 & $2\!\times\!10^{-3}$ & 0.033 & 3.531 \\
8 & $2\!\times\!10^{-3}$ & 0.033 & 3.444 \\
12 & $2\!\times\!10^{-3}$ & 0.1 & 3.413 \\
16 & $2\!\times\!10^{-3}$ & 0.1 & 3.392 \\
24 & $2\!\times\!10^{-3}$ & 0.1 & 3.372 \\
32 & $2\!\times\!10^{-3}$ & 0.1 & 3.364 \\
40 & $2\!\times\!10^{-3}$ & 0.1 & 3.361 \\
\textbf{48} & $\boldsymbol{2\!\times\!10^{-3}}$ & \textbf{0.1} & \textbf{3.360} \\
56 & $2\!\times\!10^{-3}$ & 0.1 & 3.360 \\
\bottomrule
\end{tabular}
\end{minipage}
\end{table}

\begin{table}[!htbp]
\centering\myfontsize
\captionsetup{font=footnotesize,skip=4pt}
\setlength{\tabcolsep}{2pt}
\renewcommand{\arraystretch}{1.02}
\caption{Hyperparameters for every plotted checkpoint in Figure~\ref{fig:train_validation_dynamics} (\S\ref{sec:batch_size_two_dimensional}), with the 1.5B model and a 1B-token pool. Each epoch uses its own selected LR/WD pair. All runs use the original recipe.}
\label{tab:section3_dynamics_configs}
\begin{minipage}[t]{.325\linewidth}
\centering\myfontsize
\textbf{(a) $B=16$}\par\smallskip
\begin{tabular}{@{}rrrr@{}}
\toprule
$E$ & LR & WD & CE \\
\midrule
1 & $5\!\times\!10^{-4}$ & 0.1 & 3.420 \\
2 & $5\!\times\!10^{-4}$ & 0.1 & 3.260 \\
4 & $5\!\times\!10^{-4}$ & 0.1 & 3.147 \\
8 & $5\!\times\!10^{-4}$ & 0.1 & 3.066 \\
12 & $5\!\times\!10^{-4}$ & 0.1 & 3.034 \\
16 & $5\!\times\!10^{-4}$ & 0.1 & 3.021 \\
20 & $5\!\times\!10^{-4}$ & 0.1 & 3.017 \\
\bottomrule
\end{tabular}
\end{minipage}\hfill
\begin{minipage}[t]{.325\linewidth}
\centering\myfontsize
\textbf{(b) $B=64$}\par\smallskip
\begin{tabular}{@{}rrrr@{}}
\toprule
$E$ & LR & WD & CE \\
\midrule
1 & $10^{-3}$ & 0.3 & 3.397 \\
2 & $10^{-3}$ & 0.3 & 3.230 \\
4 & $5\!\times\!10^{-4}$ & 0.3 & 3.108 \\
8 & $10^{-3}$ & 0.3 & 3.029 \\
12 & $10^{-3}$ & 0.3 & 2.996 \\
16 & $10^{-3}$ & 0.3 & 2.982 \\
20 & $10^{-3}$ & 0.3 & 2.977 \\
\bottomrule
\end{tabular}
\end{minipage}\hfill
\begin{minipage}[t]{.325\linewidth}
\centering\myfontsize
\textbf{(c) $B=512$}\par\smallskip
\begin{tabular}{@{}rrrr@{}}
\toprule
$E$ & LR & WD & CE \\
\midrule
1 & $10^{-3}$ & 0.1 & 4.000 \\
2 & $10^{-3}$ & 0.3 & 3.469 \\
4 & $10^{-3}$ & 0.3 & 3.224 \\
8 & $2\!\times\!10^{-3}$ & 0.3 & 3.124 \\
12 & $10^{-3}$ & 0.3 & 3.198 \\
\bottomrule
\end{tabular}
\end{minipage}
\end{table}

\begin{table}[!htbp]
\centering\myfontsize
\captionsetup{font=footnotesize,skip=4pt}
\setlength{\tabcolsep}{2pt}
\renewcommand{\arraystretch}{1.02}
\caption{Configurations for the data-loader comparison table in Figure~\ref{fig:data_loader_ablation}(a) (\S\ref{sec:repetition_aware}), with the 1.5B model and a 1B-token pool. All runs use untied embeddings and no embedding decay.}
\label{tab:section4_loader_ablation_configs}
\begin{minipage}[t]{.49\linewidth}
\centering\myfontsize
\textbf{$B=64$}\par\smallskip
\begin{tabular}{@{}lrrrr@{}}
\toprule
Loader & LR & WD & $E$ & CE \\
\midrule
\fixed & $10^{-3}$ & 0.3 & 16 & 3.011 \\
\ss & $10^{-3}$ & 0.3 & 20 & 2.977 \\
\ds & $10^{-3}$ & 0.3 & 24 & 2.971 \\
\bottomrule
\end{tabular}
\end{minipage}\hfill
\begin{minipage}[t]{.49\linewidth}
\centering\myfontsize
\textbf{$B=512$}\par\smallskip
\begin{tabular}{@{}lrrrr@{}}
\toprule
Loader & LR & WD & $E$ & CE \\
\midrule
\fixed & $10^{-3}$ & 0.3 & 8 & 3.205 \\
\ss & $2\!\times\!10^{-3}$ & 0.3 & 8 & 3.124 \\
\ds & $10^{-3}$ & 1 & 24 & 3.064 \\
\bottomrule
\end{tabular}
\end{minipage}
\end{table}

\begin{table}[!htbp]
\centering\myfontsize
\captionsetup{font=footnotesize,skip=4pt}
\setlength{\tabcolsep}{2pt}
\renewcommand{\arraystretch}{1.02}
\caption{Hyperparameters for every plotted checkpoint in Figure~\ref{fig:data_loader_ablation}(b) (\S\ref{sec:repetition_aware}), with the 1.5B model, a 1B-token pool, and $B=512$. All runs use untied embeddings and no embedding decay.}
\label{tab:section4_loader_curves}
\begin{minipage}[t]{.325\linewidth}
\centering\myfontsize
\textbf{\fixed}\par\smallskip
\begin{tabular}{@{}rrrr@{}}
\toprule
$E$ & LR & WD & CE \\
\midrule
1 & $10^{-3}$ & 0.3 & 4.000 \\
2 & $10^{-3}$ & 0.3 & 3.467 \\
4 & $10^{-3}$ & 0.3 & 3.240 \\
8 & $10^{-3}$ & 0.3 & 3.205 \\
12 & $10^{-3}$ & 0.3 & 3.248 \\
\bottomrule
\end{tabular}
\end{minipage}\hfill
\begin{minipage}[t]{.325\linewidth}
\centering\myfontsize
\textbf{\ss}\par\smallskip
\begin{tabular}{@{}rrrr@{}}
\toprule
$E$ & LR & WD & CE \\
\midrule
1 & $10^{-3}$ & 0.1 & 4.000 \\
2 & $10^{-3}$ & 0.3 & 3.469 \\
4 & $10^{-3}$ & 0.3 & 3.224 \\
8 & $2\!\times\!10^{-3}$ & 0.3 & 3.124 \\
12 & $10^{-3}$ & 0.3 & 3.198 \\
\bottomrule
\end{tabular}
\end{minipage}\hfill
\begin{minipage}[t]{.325\linewidth}
\centering\myfontsize
\textbf{\ds}\par\smallskip
\begin{tabular}{@{}rrrr@{}}
\toprule
$E$ & LR & WD & CE \\
\midrule
1 & $10^{-3}$ & 0.1 & 4.000 \\
2 & $10^{-3}$ & 0.3 & 3.468 \\
4 & $10^{-3}$ & 0.3 & 3.221 \\
8 & $10^{-3}$ & 0.3 & 3.137 \\
12 & $10^{-3}$ & 1 & 3.091 \\
16 & $10^{-3}$ & 1 & 3.069 \\
20 & $10^{-3}$ & 1 & 3.065 \\
24 & $10^{-3}$ & 1 & 3.064 \\
28 & $10^{-3}$ & 1 & 3.067 \\
\bottomrule
\end{tabular}
\end{minipage}
\end{table}

\begin{table}[!htbp]
\centering\myfontsize
\captionsetup{font=footnotesize,skip=4pt}
\setlength{\tabcolsep}{2pt}
\renewcommand{\arraystretch}{1.02}
\caption{Configurations for the weight-tying/embedding-decay comparison table in Figure~\ref{fig:weight_tying_ablation}(a) (\S\ref{sec:repetition_aware}), with the 1.5B model and a 1B-token pool.}
\label{tab:section4_ablation_configs}
\begin{minipage}[t]{.49\linewidth}
\centering\myfontsize
\textbf{$B=64$}\par\smallskip
\begin{tabular}{@{}lrrrrr@{}}
\toprule
Configuration & WD$_{\rm Emb}$ & LR & WD & $E$ & CE \\
\midrule
\ss, untied & 0 & $10^{-3}$ & 0.3 & 20 & 2.977 \\
\ds, untied & 0 & $10^{-3}$ & 0.3 & 24 & 2.971 \\
\ds, tied & 0 & $10^{-3}$ & 0.3 & 24 & 2.971 \\
\ds, untied & 0.3 & $10^{-3}$ & 0.3 & 32 & 2.951 \\
\ds, tied & 0.3 & $10^{-3}$ & 0.3 & 32 & 2.946 \\
\bottomrule
\end{tabular}
\end{minipage}\hfill
\begin{minipage}[t]{.49\linewidth}
\centering\myfontsize
\textbf{$B=512$}\par\smallskip
\begin{tabular}{@{}lrrrrr@{}}
\toprule
Configuration & WD$_{\rm Emb}$ & LR & WD & $E$ & CE \\
\midrule
\ss, untied & 0 & $2\!\times\!10^{-3}$ & 0.3 & 8 & 3.124 \\
\ds, untied & 0 & $10^{-3}$ & 1 & 24 & 3.064 \\
\ds, tied & 0 & $10^{-3}$ & 1 & 20 & 3.061 \\
\ds, untied & 1 & $10^{-3}$ & 1 & 28 & 3.050 \\
\ds, tied & 1 & $10^{-3}$ & 1 & 28 & 3.030 \\
\bottomrule
\end{tabular}
\end{minipage}
\end{table}

\begin{table}[!htbp]
\centering\myfontsize
\captionsetup{font=footnotesize,skip=4pt}
\setlength{\tabcolsep}{2pt}
\renewcommand{\arraystretch}{1.02}
\caption{Hyperparameters for every plotted checkpoint in Figure~\ref{fig:weight_tying_ablation}(b) (\S\ref{sec:repetition_aware}), with the 1.5B model, a 1B-token pool, $B=64$, and \ds.}
\label{tab:section4_tying_curves}
\begin{minipage}[t]{.325\linewidth}
\centering\myfontsize
\textbf{Untied, no embedding decay}\par\smallskip
\begin{tabular}{@{}rrrr@{}}
\toprule
$E$ & LR & WD & CE \\
\midrule
1 & $10^{-3}$ & 0.3 & 3.397 \\
2 & $10^{-3}$ & 0.3 & 3.229 \\
4 & $10^{-3}$ & 0.3 & 3.112 \\
8 & $10^{-3}$ & 0.3 & 3.027 \\
12 & $10^{-3}$ & 0.3 & 2.994 \\
16 & $10^{-3}$ & 0.3 & 2.978 \\
20 & $10^{-3}$ & 0.3 & 2.972 \\
24 & $10^{-3}$ & 0.3 & 2.971 \\
\bottomrule
\end{tabular}
\end{minipage}\hfill
\begin{minipage}[t]{.325\linewidth}
\centering\myfontsize
\textbf{Tied, no embedding decay}\par\smallskip
\begin{tabular}{@{}rrrr@{}}
\toprule
$E$ & LR & WD & CE \\
\midrule
1 & $10^{-3}$ & 0.3 & 3.376 \\
2 & $10^{-3}$ & 0.3 & 3.211 \\
4 & $10^{-3}$ & 0.3 & 3.097 \\
8 & $10^{-3}$ & 0.3 & 3.021 \\
12 & $10^{-3}$ & 0.3 & 2.993 \\
16 & $10^{-3}$ & 0.3 & 2.979 \\
20 & $10^{-3}$ & 0.3 & 2.973 \\
24 & $10^{-3}$ & 0.3 & 2.971 \\
\bottomrule
\end{tabular}
\end{minipage}\hfill
\begin{minipage}[t]{.325\linewidth}
\centering\myfontsize
\textbf{Tied, positive embedding decay}\par\smallskip
\begin{tabular}{@{}rrrr@{}}
\toprule
$E$ & LR & WD & CE \\
\midrule
1 & $10^{-3}$ & 0.3 & 3.381 \\
2 & $10^{-3}$ & 0.3 & 3.217 \\
4 & $10^{-3}$ & 0.3 & 3.102 \\
8 & $10^{-3}$ & 0.3 & 3.020 \\
12 & $10^{-3}$ & 0.3 & 2.986 \\
16 & $10^{-3}$ & 0.3 & 2.967 \\
20 & $10^{-3}$ & 0.3 & 2.956 \\
24 & $10^{-3}$ & 0.3 & 2.950 \\
28 & $10^{-3}$ & 0.3 & 2.947 \\
32 & $10^{-3}$ & 0.3 & 2.946 \\
36 & $10^{-3}$ & 0.3 & 2.946 \\
\bottomrule
\end{tabular}
\end{minipage}
\end{table}

\begin{table}[!htbp]
\centering\myfontsize
\captionsetup{font=footnotesize,skip=4pt}
\setlength{\tabcolsep}{2pt}
\renewcommand{\arraystretch}{1.02}
\caption{Hyperparameters for the repetition-tuned recipe across all model sizes and batch sizes, in Figure~\ref{fig:recipe_batch_optima} (\S\ref{sec:repetition_aware}), on a 1B-token pool. The same checkpoints were used for the validation loss and the Hellaswag BPB. Original-recipe configurations are in Table~\ref{tab:section3_selected_configs}.}
\label{tab:section4_recipe_configs}
\begin{minipage}[t]{.325\linewidth}
\centering\myfontsize
\textbf{1.5B}\par\smallskip
\begin{tabular}{@{}rrrrr@{}}
\toprule
$B$ & LR & WD & $E$ & CE \\
\midrule
64 & $10^{-3}$ & 0.3 & 32 & 2.946 \\
128 & $10^{-3}$ & 0.3 & 20 & 2.979 \\
256 & $10^{-3}$ & 1 & 40 & 3.019 \\
512 & $10^{-3}$ & 1 & 28 & 3.030 \\
\bottomrule
\end{tabular}
\end{minipage}\hfill
\begin{minipage}[t]{.325\linewidth}
\centering\myfontsize
\textbf{474M}\par\smallskip
\begin{tabular}{@{}rrrrr@{}}
\toprule
$B$ & LR & WD & $E$ & CE \\
\midrule
64 & $2\!\times\!10^{-3}$ & 0.1 & 56 & 3.064 \\
128 & $2\!\times\!10^{-3}$ & 0.3 & 80 & 3.060 \\
256 & $2\!\times\!10^{-3}$ & 0.3 & 64 & 3.064 \\
512 & $2\!\times\!10^{-3}$ & 0.3 & 40 & 3.105 \\
\bottomrule
\end{tabular}
\end{minipage}\hfill
\begin{minipage}[t]{.325\linewidth}
\centering\myfontsize
\textbf{153M}\par\smallskip
\begin{tabular}{@{}rrrrr@{}}
\toprule
$B$ & LR & WD & $E$ & CE \\
\midrule
64 & $2\!\times\!10^{-3}$ & 0.1 & 224 & 3.293 \\
128 & $2\!\times\!10^{-3}$ & 0.1 & 240 & 3.285 \\
256 & $2\!\times\!10^{-3}$ & 0.1 & 208 & 3.298 \\
512 & $2\!\times\!10^{-3}$ & 0.3 & 256 & 3.322 \\
\bottomrule
\end{tabular}
\end{minipage}
\end{table}

\begin{table}[!htbp]
\centering\myfontsize
\captionsetup{font=footnotesize,skip=4pt}
\setlength{\tabcolsep}{2pt}
\renewcommand{\arraystretch}{1.02}
\caption{Hyperparameters for every repetition-tuned point in the data-pool-size comparison of Figure~\ref{fig:pool_size_best_validation_ce} (\S\ref{sec:repetition_aware}). 
$\Du$ is the unique-data pool size in billions of tokens.}
\label{tab:section4_pool_configs}
\begin{minipage}[t]{.325\linewidth}
\centering\myfontsize
\textbf{1.5B}\par\smallskip
\begin{tabular}{@{}rrrrrr@{}}
\toprule
$\Du$ & $B$ & LR & WD & $E$ & CE \\
\midrule
0.333 & 64 & $10^{-3}$ & 1 & 28 & 3.255 \\
1 & 64 & $10^{-3}$ & 0.3 & 32 & 2.946 \\
3 & 128 & $10^{-3}$ & 0.3 & 80 & 2.767 \\
\bottomrule
\end{tabular}
\end{minipage}\hfill
\begin{minipage}[t]{.325\linewidth}
\centering\myfontsize
\textbf{474M}\par\smallskip
\begin{tabular}{@{}rrrrrr@{}}
\toprule
$\Du$ & $B$ & LR & WD & $E$ & CE \\
\midrule
0.333 & 64 & $2\!\times\!10^{-3}$ & 0.3 & 32 & 3.281 \\
1 & 128 & $2\!\times\!10^{-3}$ & 0.3 & 80 & 3.060 \\
3 & 256 & $2\!\times\!10^{-3}$ & 0.1 & 112 & 2.909 \\
\bottomrule
\end{tabular}
\end{minipage}\hfill
\begin{minipage}[t]{.325\linewidth}
\centering\myfontsize
\textbf{153M}\par\smallskip
\begin{tabular}{@{}rrrrrr@{}}
\toprule
$\Du$ & $B$ & LR & WD & $E$ & CE \\
\midrule
0.333 & 64 & $2\!\times\!10^{-3}$ & 0.3 & 144 & 3.452 \\
1 & 128 & $2\!\times\!10^{-3}$ & 0.1 & 240 & 3.285 \\
3 & 256 & $2\!\times\!10^{-3}$ & 0.1 & 640 & 3.196 \\
\bottomrule
\end{tabular}
\end{minipage}
\end{table}

\clearpage

\newpage
\section{Parallelism and Step-Time Estimation}
\label{app:parallelism-estimation}

This appendix summarizes the H100 performance model used in
\S\ref{subsec:aggressive_model_parallelism}.
We compare (1) a conventional \textbf{thoughput-first} policy that prioritizes tokens per GPU-second with (2) a \textbf{latency-first} policy that minimizes optimizer-step time.
The estimates are projections rather than measurements of every configuration.

\paragraph{Setup and policies.}
We model dense decoder-only Transformers at sequence length $S=4{,}096$, with
no context parallelism. Let $d,t,p,m,$ and $a$ denote the data-parallel,
tensor-parallel, pipeline-parallel, microbatch, and gradient-accumulation
degrees. Then
\begin{equation}
    B=dma, \qquad M=tp, \qquad G=dtp.
    \label{eq:parallelism-notation}
\end{equation}
We assume FP8 matrix multiplications, BF16 parameters and communication,
FlashAttention, activation checkpointing, sequence parallelism for $t>1$,
and distributed AdamW with sharded optimizer states. Parameters remain
replicated across DP groups; we do not substitute ZeRO-3 or FSDP for TP or PP.
Both policies fix the global batch at $B=1{,}024$ sequences and gradient
accumulation at $a=1$.

The throughput-first policy uses a microbatch near the
per-GPU throughput optimum and the smallest model-parallel degree that fits.
The latency-first policy instead fixes $m=1$, hence $d=1{,}024$, and searches
$t\in\{1,2,4,8\}$ and $p\in\{1,2,4,8\}$ subject to memory, divisibility, and
topology constraints. TP groups are kept within one eight-GPU NVLink domain.

\paragraph{Configuration-dependent compute and pipeline time.}
For parameter count $N$, the forward-and-backward work per sequence is
$F_{\rm seq}\approx6NS$. Rather than assigning one effective FLOP rate to all
latency-first configurations, let $P_{\rm eff}(N,t,p,m)$ denote the effective
per-GPU rate for the particular configuration. This rate includes TP
collectives and the loss of kernel efficiency as local matrices and pipeline
stages become smaller. For one microbatch, the pipeline critical path is
\begin{equation}
 T_{\rm FB}(t,p,m)
 =p\,T_{\rm stage}(N,t,p,m)
 \approx \frac{6NSm}{tP_{\rm eff}(N,t,p,m)}.
 \label{eq:pipeline-time-model}
\end{equation}
The second form is exact only when stages are balanced. Writing the first form
explicitly matters: PP does not ideally shorten the single-microbatch compute
path, and can lengthen it when smaller stages lower utilization.

Forward activations and backward activation gradients cross each of the
$p-1$ pipeline boundaries. For hidden dimension $h$, their approximate BF16
volume is
\begin{equation}
 V_{\rm PP}\approx\frac{4(p-1)mSh}{t}, \qquad
 T_{\rm PP}=\frac{V_{\rm PP}}{\beta_{\rm PP}}
             +2(p-1)\alpha_{\rm PP}.
 \label{eq:pp-time-model}
\end{equation}
This term is normally much smaller than DP gradient communication, but it is
monotone in $p$ and therefore prevents cost-free pipeline splitting.

\paragraph{Data-parallel communication and overlap.}
With BF16 communication, a distributed-optimizer step performs a gradient
reduce-scatter and a parameter all-gather. Their per-rank volumes are
\begin{equation}
 V_{\rm RS}\approx V_{\rm AG}\approx
 \frac{2N}{tp}\frac{d-1}{d}.
 \label{eq:dp-volume-model}
\end{equation}
Let $T_{\rm RS}$ and $T_{\rm AG}$ include both bandwidth and collective
latency. Gradient reduction can overlap only with the remaining backward
schedule after a bucket becomes ready; parameter gathering can overlap only
with optimizer work and the appropriate portion of the next forward pass.
For a compact fitted model, we combine the phases into
$T_{\rm DP}=T_{\rm RS}+T_{\rm AG}$ and expose only the portion that cannot
overlap the forward--backward critical path:
\begin{equation}
 T_{\rm DP}^{\rm exp}
 =T_{\rm DP}-\eta\min(T_{\rm FB},T_{\rm DP}),
 \label{eq:exposed-dp-model}
\end{equation}
where $\eta$ is a fitted overlap fraction. This interpolates between no
overlap ($\eta=0$) and the optimistic maximum model ($\eta=1$) without
allowing communication to overlap more compute than is available.

Our steady-state step estimate is
\begin{equation}
 \widehat T_{\rm step}(d,t,p,m)
 =T_{\rm FB}(t,p,m)+T_{\rm DP}^{\rm exp}(d,t,p)
  +T_{\rm PP}(t,p,m)+T_{\rm opt}(t,p)+T_0,
 \label{eq:step-time-estimate}
\end{equation}
where $T_{\rm opt}$ and $T_0$ cover optimizer and fixed framework costs. All
terms that vary with $t$ or $p$ are kept explicit or calibrated separately;
they are not absorbed into a configuration-independent residual.

\paragraph{Calibration and selection.}
We anchor H100 compute rates and feasible configurations to NVIDIA Megatron
measurements \citep{nvidia_megatron_performance,nvidia_activation_recompute}.
For 70B we use $P_{\rm eff}=0.65$ PFLOP/s/GPU at TP8, $\eta=0.5$, and fit the effective DP bandwidth to the stated TP4/PP1 latency of 2.10 s. This gives
$\beta_{\rm DP}=40$ GB/s. We apply stage-efficiency penalties of 0\%, 3\%,
and 8\% at PP2, PP4, and PP8, respectively. The fitted combined values of
$T_{\rm PP}+T_{\rm opt}+T_0$ are 0.05, 0.09, 0.11, and 0.18 s for
PP1, PP2, PP4, and PP8. These coefficients are held fixed while evaluating
the candidates; they are not refit to favor a particular configuration.

The resulting 70B candidate latencies are
\begin{center}
\begin{tabular}{lrrrrrrrr}
\toprule
$(t,p)$ &(4,1)&(4,2)&(4,4)&(4,8)&(8,1)&(8,2)&(8,4)&(8,8)\\
\midrule
$\widehat T_{\rm step}$ (s)
&2.10&1.28&1.01&1.01&1.08&0.69&\textbf{0.56}&0.59\\
\bottomrule
\end{tabular}
\end{center}
Thus the fitted latency objective selects TP8/PP4. PP8 provides no ideal
single-microbatch compute reduction after PP4 reaches the compute floor, and
its lower stage efficiency and additional PP overhead make it slower.

\end{document}